%% file: neurips_2026.tex
\documentclass{article}

\PassOptionsToPackage{numbers}{natbib}                                      
\usepackage[preprint]{neurips_2026}
\usepackage{graphicx}
\usepackage{booktabs}
\usepackage{enumitem}
\usepackage{wrapfig}    
\usepackage{float} 
\usepackage{multirow}
\usepackage{amsmath}  
\usepackage{amssymb} 
\usepackage{algorithm}
\usepackage{algpseudocode}

\usepackage[utf8]{inputenc} 
\usepackage[T1]{fontenc}    
\usepackage{hyperref}       
\usepackage{url}            
\usepackage{booktabs}       
\usepackage{amsfonts}       
\usepackage{pifont}
\usepackage{nicefrac}       
\usepackage{microtype}      
\usepackage{xcolor}         
\newcommand{\hlfirst}[1]{\colorbox[HTML]{6FA8DC}{#1}}  
\newcommand{\hlsecond}[1]{\colorbox[HTML]{CFE2FF}{#1}}  
\title{ARC-STAR: Auditable Post-Hoc Correction for PDE Foundation Models}

\author{%
{\bfseries Chengze Li$^{1,*}$ \quad Lingwei Wei$^{1}$ \quad Li Sun$^{2}$ \quad Hongbo Lv$^{3}$ \quad Jie Yang$^{1}$}\\
{\bfseries Hanrong Zhang$^{1}$ \quad Kening Zheng$^{1}$ \quad Wei-Chieh Huang$^{1}$ \quad Enze Ma$^{1}$ \quad Philip S. Yu$^{1}$}\\
{\normalfont $^{1}$University of Illinois Chicago}\\
{\normalfont $^{2}$Beijing University of Posts and Telecommunications}\\
{\normalfont $^{3}$North China Electric Power University}\\
{\normalfont $^{*}$Correspondence: \texttt{chengze6@uic.edu}}}

\begin{document}

\maketitle

\begin{abstract}
Partial differential equation (PDE) foundation models are pretrained networks that forecast how physical fields like velocity and pressure evolve from a single reusable solver. On unfamiliar flows their predictions drift step by step, errors concentrate in a few regions, yet retraining destabilizes the network and uniform post-hoc correction overlooks this spatial concentration. To address this, we propose a frozen-solver post-hoc correction framework, \textbf{A}daptive \textbf{R}isk-\textbf{C}alibrated \textbf{S}patial \textbf{T}riage for \textbf{A}uditable \textbf{R}efinement (ARC-STAR). ARC-STAR organizes correction into three stages: a global corrector removes broad solver bias, a blockwise local refiner cleans the post-global residual, and, at deployment, a label-free score routes refinement to high-risk blocks under a compute budget. The framework is designed to be (i) frozen-host, preserving the pretrained solver without fine-tuning; (ii) auditable, with global and local stages trained and evaluated separately for measurable contributions; and (iii) budget-aware, using a blockwise interface that either refines the full field or routes limited compute to high-risk regions. Across five flow benchmarks spanning ten regime cells, ARC-STAR is the only method that cuts velocity rollout error by at least 36$\times$ over raw Poseidon on every cell. The global stage reduces raw host error by 91--99$\%$, and the local stage further reduces the remaining post-global residual by up to 94.4$\%$. Our code implementation is available at {\url{https://anonymous.4open.science/r/arc_star}}.

\end{abstract}

\section{Introduction}
\label{sec:intro}
A PDE foundation model is a neural network pretrained on many partial differential equations~\citep{herde2024poseidon, hao2024dpot, mccabe2023mpp} and reused as a frozen, general-purpose
solver at deployment, with examples such as Poseidon, DPOT, PROSE-FD, and MPP~\citep{herde2024poseidon, hao2024dpot, liu2024prosefd, mccabe2023mpp, serrano2025zebra}. When the deployment regime differs from pretraining,
the autoregressive rollout feeds each prediction back as the next input and amplifies even small one-step errors over many time steps~\citep{lippe2023pderefiner, kochkov2021machine}. Updating the
network for each new regime is unstable and expensive, so we ask a simpler question with the network held fixed, in the spirit of test-time adaptation in vision~\citep{wang2021tent} and language: how can a trained correction layer be both (i) decomposed into auditable stages, and (ii) deployed at a tunable compute budget without retraining?

\begin{figure}[!t]
\centering
\includegraphics[width=0.91\linewidth]{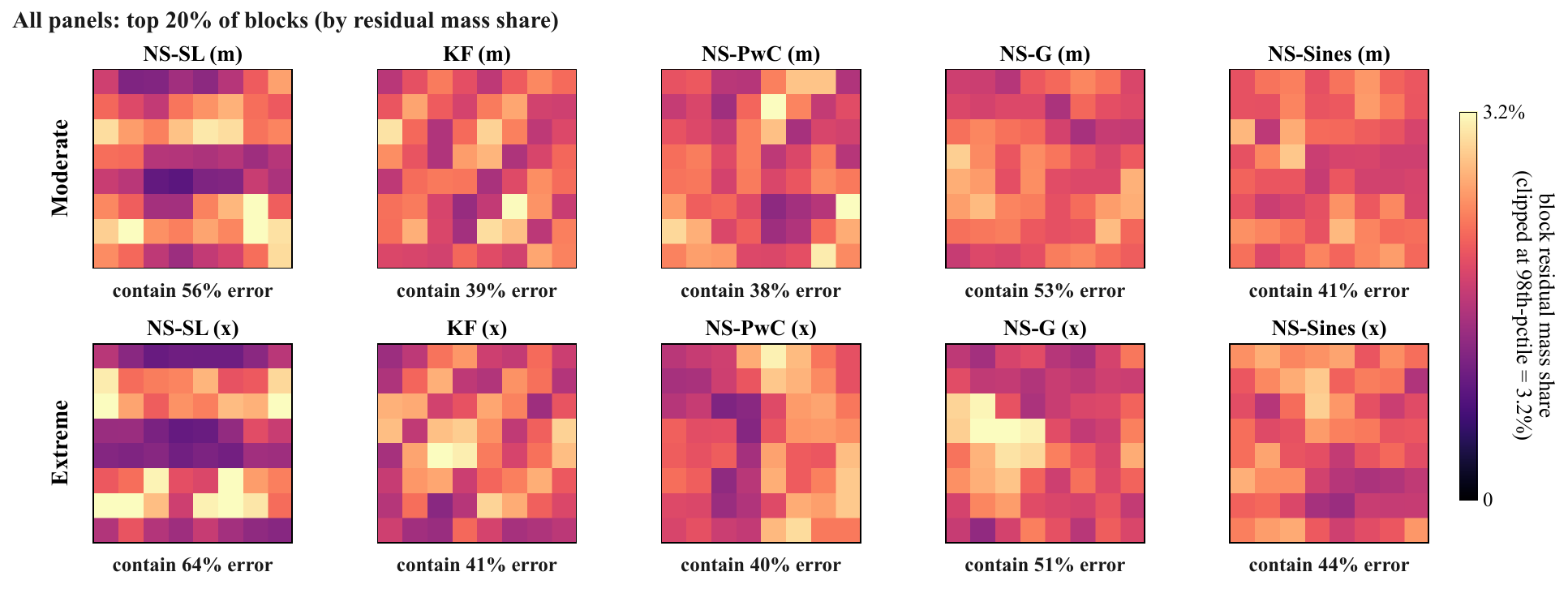}
\caption{\textbf{Post-broad-correction residual is spatially concentrated, not uniform.} Each cell is partitioned into 64 blocks of $16{\times}16$ pixels; tile color is per-block error share, with per-panel annotation reporting the share carried by the highest-error 20$\%$ of blocks (mean per-cell Gini coefficient 0.48, between the uniform null and the sparse-activation regime). Block ranking is an offline diagnostic; the deployed selector is label-free.}
\label{fig:intro_motivation}
\end{figure}

To address rollout error on a pretrained host, three families of methods have emerged: host-modifying fine-tuning of the pretrained network, dense post-processing~\citep{lippe2023pderefiner, stachenfeld2022learned,
kochkov2021machine, um2020solverintheloop} that stacks a learned correction layer over every spatial location, and hand-crafted spatial indicators such as vorticity, signal-domain
detail~\citep{donoho1994ideal, mallat1999wavelet}, or test-time uncertainty~\citep{gal2016dropout, lakshminarayanan2017simple} that flag where to spend more compute. The dominant frozen-host route is
dense post-processing, which rests on an implicit assumption: that the residual after broad correction is spatially uniform, so that compute spent uniformly across every pixel is well-spent. However,
whether this uniformity assumption holds on rollouts of pretrained PDE solvers has not been quantitatively tested across regimes. To investigate, we apply a broad residual correction to a frozen
Poseidon host's prediction on ten benchmark--regime cells, partition each predicted field into 64 non-overlapping blocks of $16{\times}16$ pixels, and rank the blocks by their share of remaining error.
Fig.~\ref{fig:intro_motivation} reports per-block error shares across all ten cells, highlighting~\mbox{two key observations:}

\begin{itemize}[leftmargin=*]
    \item[\ding{182}]\label{phe:1} \textbf{\textit{Broad correction does not homogenize the residual.}}
    A corrector with full-field receptive width has access to every pixel and could in principle spread its capacity uniformly. However, the post-correction residual remains non-trivially concentrated: a small fraction of blocks continues to dominate the rollout loss, and the per-cell concentration sits well above the uniform null while remaining clearly below the regime of sparse activations. This pattern holds with comparable strength across all five flow families and both viscosity regimes, suggesting that the structured component of rollout error survives full-field cleanup rather than being smoothed into noise.

    \item[\ding{183}]\label{phe:2} \textbf{\textit{The concentration tracks a known turbulent mechanism rather than statistical noise.}}
    The spatial pattern mirrors a well-documented structure of turbulent flows, where error concentrates in vortex cores and shear layers~\citep{pope2000turbulent}. The same concentration appears on a structurally distinct DPOT-Ti backbone (Appendix~\ref{app:cross-host}), indicating that the mechanism is host-independent rather than an artifact of Poseidon's pretraining. This decoupling licenses a single hand-designed routing rule to transfer across the cell suite without per-cell tuning, requiring no host-specific calibration at deployment.
\end{itemize}

These observations rule out dense uniform correction as compute-efficient and expose a fundamental compute-allocation gap: under concentration this strong, processing every pixel with the same operator
wastes most of the budget on regions that already contain little remaining error. They also point toward a specific design. A correction layer that operates one block at a time, in the spirit of
adaptive mesh refinement in classical numerics~\citep{berger1984adaptive, dorfler1996convergent}, paired with a label-free rule that decides which blocks to act on at
deployment~\citep{graves2017adaptive, shazeer2017outrageously}, can spend a fixed compute budget where the residual actually lives. We turn next to a specific architecture that realizes these requirements on a frozen host.

We propose \textbf{ARC-STAR}, a deployment-time correction layer that wraps a frozen PDE foundation model with two trainable stages and a label-free spatial selector. The first stage broadly corrects the predicted field. The second stage processes one block at a time: it reads a small surrounding neighborhood as context but writes only that block, and the same trained module operates either on every block or on the highest-risk blocks under a compute budget. The two stages are trained and evaluated separately, so the share of error each removes can be read off as two distinct numbers, providing a stage-wise diagnostic that separates a broad-correction shortfall from a local-correction shortfall on each new regime encountered at deployment, without host retraining.

Our contributions are as follows:
\begin{itemize}
    \item \textbf{Empirical finding.} The residual after a broad correction stage is spatially concentrated on every one of ten benchmark--regime cells from a frozen Poseidon host, with the top 20\% of blocks carrying 38\% to 64\% of remaining error.
    \item \textbf{ARC-STAR framework.} A frozen-host correction architecture (Sec.~\ref{sec:method}) whose broad and local stages are trained sequentially, supporting per-stage empirical attribution.
    \item \textbf{Benchmark performance.} ARC-STAR is best in seven of ten regime cells, outperforms two parameter-efficient Poseidon fine-tuning baselines on at least nine of ten cells, and remains competitive against nine external routing policies under shared compute budgets.
\end{itemize}

\section{Related Work}

\paragraph{Neural operators and PDE foundation models.}
Operator learning treats PDE solving as learning maps between function spaces rather than fixed-dimensional vectors~\citep{lu2021deeponet, li2021fourier, kovachki2023neuraloperator}. This line
has produced spectral, multiscale, graph, geometry-aware, attention-based, convolutional, and tensorized spectral variants~\citep{gupta2021multiwavelet, rahman2023uno, pfaff2021meshgraphnets,
sanchezgonzalez2020learning, brandstetter2022mppde, li2023geofno, cao2021choose, raonic2023cno, kossaifi2024tfno}, evaluated on community benchmarks~\citep{takamoto2022pdebench, gupta2023pdearena}.
More recently, pretrained PDE foundation models including Poseidon, DPOT, PROSE-FD, MPP, Walrus, MORPH, and Zebra extend operator learning toward reusable solvers across heterogeneous dynamics and transfer settings~\citep{herde2024poseidon, hao2024dpot, liu2024prosefd, mccabe2023mpp, mccabe2025walrus, rautela2025morph, serrano2025zebra}, while weather-scale forecasters extend this paradigm to global atmospheric
prediction~\citep{pathak2022fourcastnet, lam2023graphcast}. ARC-STAR is complementary to this trend: it does not propose a new host architecture or pretraining strategy, but instead studies
how to correct a frozen pretrained host at deployment time without further pretraining or task-specific fine-tuning.

\textbf{Rollout correction, refinement, and host-side adaptation.} A separate line addresses long-horizon degradation through refinement or adaptation: iterative neural refinement performs dense denoising during rollout~\citep{lippe2023pderefiner}; learned post-processing and hybrid solver corrections augment base simulators with residual modules~\citep{kochkov2021machine,stachenfeld2022learned,um2020solverintheloop,bar2019learning}; physics-informed methods inject PDE structure into training or test-time optimization~\citep{raissi2019physics,li2021pino,shu2023physicsinformed}; classical projection enforces incompressibility deterministically~\citep{chorin1968numerical,temam1977navier}; and PITA aligns rollout states inside the host via self-supervision~\citep{zhu2025pita}. \citet{wei2025inc} prove that direct additive correction can amplify per-step errors over autoregressive rollouts; concurrent host-side work modifies training (spatially-adaptive Jacobian regularization, JAWS~\citep{jaws2026}) or rollout state (EMA residual-monitored solver interventions, ANCHOR~\citep{anchor2025}). ARC-STAR instead keeps $H$ frozen, audits broad and local cleanup as two stages, and applies a halo-read, center-write block that leaves unselected regions exactly equal to the post-global prediction; empirically, error growth remains bounded across all in-distribution rollout horizons admitted by the Poseidon test split.

\paragraph{Adaptive spatial allocation and routing.}
The idea of non-uniform computation is classical in numerical PDEs, especially in adaptive mesh refinement (AMR) and a posteriori marking~\citep{berger1984adaptive,
dorfler1996convergent}. In machine learning, related selection signals span uncertainty and ensembles~\citep{gal2016dropout, lakshminarayanan2017simple, ayhan2018test, seung1992query}, novelty and
OOD scores~\citep{lee2018simple, sun2021react, sun2022knn}, saliency and gradient attribution~\citep{simonyan2014deep, selvaraju2017gradcam}, signal-domain detail~\citep{donoho1994ideal,
mallat1999wavelet}, consistency self-ensembling~\citep{laine2017temporal, tarvainen2017mean, zhu2017cyclegan}, conformal risk control~\citep{angelopoulos2023conformal}, adaptive token
selection~\citep{ryoo2021tokenlearner, rao2021dynamicvit, graves2017adaptive}, and mixture-of-experts routing~\citep{shazeer2017outrageously, fedus2022switch}. ARC-STAR differs in two ways.
First, the routed unit is a spatial block aligned with the local refiner's halo-read, center-write contract, not a generic token or example. Second, all candidate routing rules are compared
head-to-head with the same frozen host, the same global and local stages, and the same compute budget grid (Sec.~\ref{sec:exp_routing_qual}), thereby isolating block-selection as the only varying
factor in a deployment-faithful setting.


\section{Methodology}
\label{sec:method}

\subsection{Problem setting and overview}
\label{sec:method_overview}

We study deployment-time correction for a frozen pretrained PDE foundation model. Let $x_t \in \mathbb{R}^{C \times H \times W}$ denote the state at time $t$, and let $x^*_{t+1}$ denote the
ground-truth next state, used only at training and evaluation. The host $H$ produces a one-step forecast $\hat{x}_{t+1} = H(x_t)$, and autoregressive rollout feeds the corrected prediction back as the
next input. The host parameters remain fixed throughout. We write $\Pi_{uv}$ for the projection onto the velocity channels; all learned residuals and reported losses are applied through $\Pi_{uv}$,
while non-velocity channels pass through unchanged.

\begin{figure}[t]
    \centering
    \includegraphics[width=\linewidth]{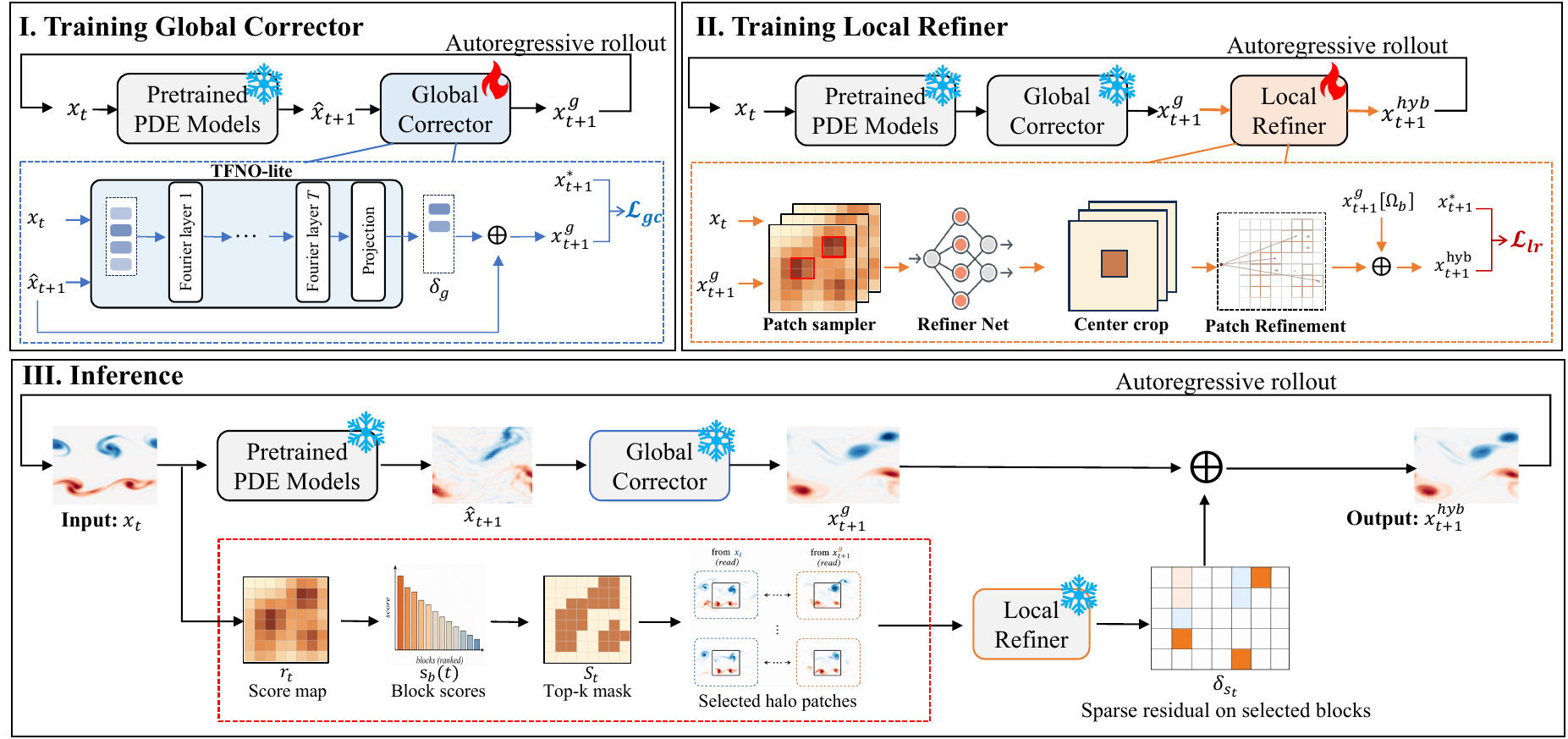}
    \caption{
\textbf{Overview of ARC-STAR.}
\textbf{I. Global-stage training:} the frozen host stays fixed while a full-field residual operator learns a residual correction toward the ground-truth next state.
\textbf{II. Local-stage training:} after the global stage is fixed, a blockwise local refiner is trained on post-global fields with a halo-read, center-write patch interface, matching the inputs seen at deployment.
\textbf{III. Inference:} a hand-designed, label-free risk score selects the active blocks, and the local refiner runs densely or under a top-$k$ budget.
The staged design makes global and local stages separately auditable.
}
    \label{fig:method_overview}
\end{figure}

ARC-STAR wraps this frozen host in three stages mirroring the three panels of Fig.~\ref{fig:method_overview}. \textbf{Stage~I} (Fig.~\ref{fig:method_overview}.I) trains a global corrector $G_\phi$ to remove broad field-level bias (Sec.~\ref{sec:method_global}). \textbf{Stage~II} (Fig.~\ref{fig:method_overview}.II) trains a blockwise local refiner $L_\theta$ on post-global residuals under a halo-read, center-write contract; Algorithm~\ref{alg:training} realizes this training in two steps, a dense patch-pretraining step (Step~2a) followed by an autoregressive fine-tuning step with $k{=}B$ (Step~2b), keeping the local module independent of any deployment budget and reusable across budget levels without retraining (Sec.~\ref{sec:method_local}). \textbf{Stage~III} (Fig.~\ref{fig:method_overview}.III) runs at deployment: a label-free score ranks blocks and routes either the full set or a top-$k$ subset to $L_\theta$ under the same halo-read, center-write contract (Sec.~\ref{sec:method_inference}). Throughout, $\{\Omega_b\}_{b=1}^{B}$ denotes a partition of the spatial domain into non-overlapping center blocks, and $\Omega_b^{+h}$ denotes the same block enlarged by a halo of width $h$. The local refiner reads $\Omega_b^{+h}$ but writes only $\Omega_b$. This contract is the key to both dense and budgeted inference: the same trained local module refines all blocks when $k{=}B$ and only a selected subset when $k{<}B$, without retraining.

\subsection{Training the global corrector}
\label{sec:method_global}

The global stage performs broad cleanup before any local routing is introduced. Given the current state and the raw host forecast, the global corrector predicts a velocity-channel residual that is added back to $\hat{x}_{t+1}$ to produce the post-global one-step state:
\begin{equation}
    x^g_{t+1} = \hat{x}_{t+1} + G_\phi(x_t, \hat{x}_{t+1}),
    \label{eq:global}
\end{equation}
where $G_\phi$ writes only the velocity channels $(u,v)$; non-velocity channels $(I-\Pi_{uv})\hat{x}_{t+1}$ pass through unchanged at every autoregressive step.

We train $G_\phi$ autoregressively with the frozen host in the loop. Starting from $z_t = x_t$, we roll out
\begin{equation}
    \hat{z}_{t+i} = H(z_{t+i-1}),
\qquad
z^{g}_{t+i} = \hat{z}_{t+i} + \Pi_{uv}\!\left[\,G_\phi(z_{t+i-1}, \hat{z}_{t+i})\right],
\end{equation}
for $i=1,\ldots,N$, and optimize the multi-step velocity loss
\begin{equation}
    \mathcal{L}_{\mathrm{gc}}(\phi)
    \;=\;
    \sum_{i=1}^{N} \left\| \Pi_{uv}\!\left(x^{*}_{t+i} - z^{g}_{t+i}\right) \right\|_2^2.
    \label{eq:lgc}
\end{equation}
After convergence, $G_\phi$ is frozen. All subsequent local-refiner training operates on the post-global fields produced by the fixed pair $(H, G_\phi)$, so that the local stage sees the same input distribution at training and at deployment, eliminating any train-test distribution gap.

\subsection{Training the local refiner}
\label{sec:method_local}

The local refiner is trained after the global corrector is fixed and operates on post-global fields produced by the frozen pair $(H,G_\phi)$. For each block $b$, we extract the halo-extended window
from the concatenation of the current state and the post-global forecast, and predict the velocity residual,
\begin{equation}
    \Pi_{uv}\!\left(x^{\mathrm{hyb}}_{t+1}\right)\!\restriction_{\Omega_b}
    = \Pi_{uv}\!\left(x^{g}_{t+1}\right)\!\restriction_{\Omega_b}
    + w_{\mathrm{H}} \odot \delta_b, \qquad b \in \mathcal{S}_t,
    \label{eq:local}
\end{equation}
where $w_{\mathrm{H}}$ is an outer-product Hann window of side $|\Omega_b|^{1/2}$ that suppresses block-boundary discontinuities (Appendix~\ref{app:arcstar-architecture}). Unselected blocks satisfy $x^{\mathrm{hyb}}_{t+1}\!\restriction_{\Omega_b} \equiv x^{g}_{t+1}\!\restriction_{\Omega_b}$ bit-exactly. We instantiate $L_\theta$ as a ConvNeXt-style patch refiner~\citep{liu2022convnext} with $16{\times}16$ center blocks and halo width $h{=}8$ (sensitivity in Appendix~\ref{app:block-halo-sweep}). Appendix~\ref{app:hann-ablation} ablates the Hann window and shows that removing it makes every cell 7--682$\times$ worse than the raw host (median UV-$L^2$ ratio without Hann divided by raw), with degradation inversely correlated with residual smoothness.

\begin{wrapfigure}[22]{r}{0.50\linewidth}
\vspace{-2.25em}
\begin{minipage}{\linewidth}
\begin{algorithm}[H]
\caption{ARC-STAR training procedure.}
\label{alg:training}
\scriptsize
\begin{algorithmic}[1]
\setlength{\itemsep}{0pt}
\Require trajectories $\mathcal{D}$;\ AR length $N$;\ partition $\{\Omega_b\}_{b=1}^{B}$.
\Statex \textit{\textbf{Step 1: train global corrector $G_\phi$ (Fig.~\ref{fig:method_overview}.I).}}
\For{batch $(x_t, x^*_{t+1{:}t+N}) \!\sim\! \mathcal{D}$}
  \State $z_t \gets x_t$ \Comment{init AR state}
  \For{$i=1,\ldots,N$}
    \State $\hat{z}_{t+i}\!\gets\! H(z_{t+i-1})$
    \State $z^g_{t+i}\!\gets\!\hat{z}_{t+i}\!+\!\Pi_{uv}[G_\phi(z_{t+i-1},\hat{z}_{t+i})]$
    \State $z_{t+i}\!\gets\! z^g_{t+i}$ \Comment{closed-loop feedback}
  \EndFor
  \State minimize $\sum_i\|\Pi_{uv}(x^*_{t+i}{-}z^g_{t+i})\|^2$ w.r.t.\ $\phi$
\EndFor;\ \textbf{freeze} $G_\phi$.
\Statex \textit{\textbf{Step 2a: dense patch pretraining of $L_\theta$ (Fig.~\ref{fig:method_overview}.II).}}
\For{$(x_t, x^*_{t+1})\!\sim\!\mathcal{D}$, every $b\!\in\!\{1,\ldots,B\}$}
  \State $W_b\!\gets\!\mathrm{halo\_extract}(x_t,x^g_{t+1};\Omega_b^{+h})$
  \State $\delta_b\!\gets\!\mathcal{C}[L_\theta(W_b)]$
\EndFor
\State minimize $\sum_b w_b\|\Pi_{uv}(x^*_{t+1}{-}x^g_{t+1})|_{\Omega_b}{-}\delta_b\|^2$ w.r.t.\ $\theta$
 \Statex \textit{\textbf{Step 2b: dense AR fine-tuning of $L_\theta$, $k{=}B$ (Fig.~\ref{fig:method_overview}.II).}}
\For{batch $\!\sim\!\mathcal{D}$, AR step $i\!=\!1,\ldots,N$}
  \State apply $L_\theta$ on every block via Eq.~\eqref{eq:local} $\Rightarrow x^{\mathrm{hyb}}_{t+i}$
\EndFor
\State minimize $\sum_i \lVert \Pi_{uv}(x^*_{t+i} - x^{\mathrm{hyb}}_{t+i}) \rVert^2 + \lambda_{\mathrm{wd}} \lVert \theta \rVert_2^2$ w.r.t.\ $\theta$ \Comment{$\lambda_{\mathrm{wd}}=10^{-4}$ via AdamW (Table~\ref{tab:training-stage2})}
\State \Return $G_\phi,\ L_\theta$
\end{algorithmic}
\end{algorithm}
\end{minipage}
\vspace{-1.0em}
\end{wrapfigure}

Algorithm~\ref{alg:training} lays out the training protocol; the prose below mirrors the steps in the box. \textbf{Step~1} trains the global corrector $G_\phi$ in the AR loop with the host $H$ frozen, then freezes $G_\phi$ before any local-refiner training begins; this realizes Fig.~\ref{fig:method_overview}.I. \textbf{Step~2a} densely pretrains $L_\theta$ on every block of the post-global field with an optional per-patch reweighting $w_b\!\geq\!0$ (full form in Appendix~\ref{app:arcstar-training}; $w_b\!\equiv\!1$ recovers the uniform variant); this phase teaches the refiner the blockwise residual geometry it will see at deployment, before any rollout interaction is introduced. \textbf{Step~2b} places the pretrained refiner back into the full hybrid rollout with $k{=}B$, refining every block at every AR step under the halo-read center-write rule of Eq.~\eqref{eq:local}. Steps~2a and~2b together realize Fig.~\ref{fig:method_overview}.II. Routing is introduced only at inference (Sec.~\ref{sec:method_inference}, realizing Fig.~\ref{fig:method_overview}.III); decoupling Step~2b from the routing decision exploits the patch-locality of $L_\theta$: each forward pass takes a single halo window unaware of other blocks' refinement status, so the same $L_\theta$ trained at $k=B$ applies exactly to any subset $\mathcal{S}_t(k)$ at deployment, with unrefined blocks bit-exactly equal to $x^g_{t+1}$ (Eq.~\eqref{eq:local}). The frontier curves in Fig.~\ref{fig:routing_frontier} are smooth and monotone in $k/B$ with no kink at $k/B=1$, confirming no train-test gap across cells and budgets, as expected from the patch-locality argument above.

\subsection{Inference and stage-wise audit}
\label{sec:method_inference}

At deployment, the frozen host first produces $\hat{x}_{t+1}$ and the global corrector forms $x^{g}_{t+1}$ via Eq.~\eqref{eq:global}. We then compute a deployment-time label-free, purely forward score map combining
velocity innovation with post-global kinetic-energy gradient,
\begin{equation}
    r_t(p)\;=\;|\Delta u_t(p)| + |\Delta v_t(p)| \;+\; \lambda_{\mathrm{KE}}\!\left(|\partial_x K^{g}_{t+1}(p)| + |\partial_y K^{g}_{t+1}(p)|\right),
    \label{eq:risk}
\end{equation}
where $\Delta u_t = u^{g}_{t+1} - u_t$ and $\Delta v_t$ analogously, $K^{g}_{t+1} = \tfrac{1}{2}((u^{g}_{t+1})^2 + (v^{g}_{t+1})^2)$, and $\lambda_{\mathrm{KE}} = 0.05$ is a fixed score weight; at fixed budget $k/B=0.25$, the post-correction median ratio varies by less than $15\%$ across $\{0, 0.025, 0.05, 0.1, 0.2\}$ on all five benchmarks (Appendix~\ref{app:lambda-sweep}). The score is aggregated to per-block scores and selects the active set:
\begin{equation}
    s_b(t) \;=\; \frac{1}{|\Omega_b|}\sum_{p\in\Omega_b} r_t(p),
    \qquad
    \mathcal{S}_t(k) \;=\;
    \begin{cases}
        \{1,\ldots,B\}, & k = B \quad (\text{dense}), \\
        \mathrm{TopK}\!\big(\{s_b(t)\}_{b=1}^{B},\,k\big), & k < B \quad (\text{budgeted}).
    \end{cases}
    \label{eq:score}
\end{equation}
The local refiner applies Eq.~\eqref{eq:local} only on $\mathcal{S}_t(k)$, and the hybrid prediction is fed back autoregressively:
\begin{equation}
    x_t \;\xrightarrow{\;H\;}\; \hat{x}_{t+1}
    \;\xrightarrow{\;G_\phi\;}\; x^{g}_{t+1}
    \;\xrightarrow{\;L_\theta,\,\mathcal{S}_t(k)\;}\; x^{\mathrm{hyb}}_{t+1}
    \;=:\; x_{t+1}.
    \label{eq:rollout}
\end{equation}
The risk-aware label refers to the empirical alignment between $s_b(t)$ and realized per-block residual evaluated through the routing frontier (Sec.~\ref{sec:exp_routing_qual}); we make no formal conformal-calibration claim. Full inference pseudocode is in Appendix~\ref{app:inference}.

\paragraph{Stage-wise diagnostic decomposition.}
The two-stage structure lets us measure exactly where the improvement comes from. Let $L^{\mathrm{raw}}$, $L^{\mathrm{glob}}$, and $L^{\mathrm{hyb}}$ denote the rollout errors of the raw host, the
global-only system, and the final hybrid system. We define
\begin{equation}
    A = 1-\frac{L^{\mathrm{glob}}}{L^{\mathrm{raw}}},
    \qquad
    J_{\mathrm{loc}} = 1-\frac{L^{\mathrm{hyb}}}{L^{\mathrm{glob}}},
    \label{eq:audit_terms}
\end{equation}
so that the total realized improvement decomposes exactly as
\begin{equation}
    1-\frac{L^{\mathrm{hyb}}}{L^{\mathrm{raw}}}
    \;=\;
    A + (1-A)J_{\mathrm{loc}}.
    \label{eq:audit}
\end{equation}
This is an algebraic identity on aggregated losses, used as a bookkeeping diagnostic rather than a theoretical claim. Here $A$ is the share of raw host error already removed by the global stage, and $J_{\mathrm{loc}}$ is the share of the remaining post-global error that the local stage further removes. When ARC-STAR
underperforms, this split tells us which component limits the system: small $A$ means little global headroom, small $J_{\mathrm{loc}}$ means the residual is locally hard to clean up; and small overall improvement
with both terms healthy means the residual pattern itself does not reward block-level refinement.

\section{Experiments}
\label{sec:experiments}

In this section, we conduct extensive experiments to answer the following research questions: (\textbf{RQ1}) Does \textsc{ARC-STAR} improve 10-step rollout accuracy against dense solvers and frozen-host post-hoc references? (\textbf{RQ2}) How is the realized improvement distributed between the global and local stages? (\textbf{RQ3}) Under reduced local coverage, does the label-free score select the useful blocks? (\textbf{RQ4}) Does the same recipe transfer to a structurally different frozen host?

\subsection{Experimental Settings}
\label{sec:exp_setup}

\textbf{Host, data, and metric.} We use Poseidon~\citep{herde2024poseidon} as the frozen host on five incompressible-flow families, NS-SL, KF, NS-PwC, NS-G, and NS-Sines, each under moderate (m) and extreme (x) regimes, for ten benchmark--regime cells in total. The reported metric is the per-run median 10-step UV relative-$L^2$ divided by raw Poseidon on the same run, so raw Poseidon equals 1.000 on every cell and lower is better. Each cell uses four held-out trajectories with $t_0\!\in\!\{5,10\}$, giving $n=8$ paired rollouts. Splits, bootstrap intervals, absolute $L^{\mathrm{raw}}$ values, and physical-fidelity diagnostics are in Appendices~\ref{app:benchmarks}, \ref{app:metric-stats}, \ref{app:absolute-raw}, and~\ref{app:divergence}. Divergence is reported only to confirm that velocity-only correction does not introduce spurious incompressibility violation, and is not used to select cells or methods.

\textbf{ARC-STAR recipe.} The deployed system uses a TFNO global corrector ($6$ Fourier layers, $d_v{=}64$, $m{=}16$ modes) and a ConvNeXt-style local refiner (width $96$, depth $6$, $16{\times}16$ blocks, halo $h{=}8$), with full local coverage ($k/B{=}1$); the same trained $L_\theta$ is reused at every routing budget in Fig.~\ref{fig:routing_frontier}. Training follows Algorithm~\ref{alg:training} sequentially: $G_\phi$ via Eq.~\eqref{eq:lgc}, then $L_\theta$ via dense patch pretraining followed by AR fine-tuning at $k{=}B$. Cross-family baselines share the same $200$ training trajectories per cell and the same 5-step closed-loop loss as $G_\phi$; the $10$-step rollout is evaluation-only (Appendix~\ref{app:baselines}).

\textbf{Baselines.} External baselines span four families: \emph{Dense neural operators} (FNO~\citep{li2021fourier}, TFNO~\citep{kossaifi2024tfno}, UNO~\citep{rahman2023uno}, U-Net~\citep{ronneberger2015unet}, ResNet~\citep{he2016resnet}, PDEArena~\citep{gupta2023pdearena}); \emph{iterative refinement} (PDE-Refiner~\citep{lippe2023pderefiner}); \emph{post-hoc correction} (Leray projection~\citep{chorin1968numerical}, DenseRes, Partial-FT~\citep{herde2024poseidon}, LoRA-FT~\citep{hu2022lora}); and \emph{test-time correction} (PINN-Residual TTO~\citep{raissi2019physics}, and PINO-TTO-lite~\citep{li2021pino}). DenseRes, PINO-TTO-lite, and the learned-error head (Fig.~\ref{fig:routing_frontier}) are implementation-controlled references trained on identical trajectories under matched parameter budgets; full architectures and training details in Appendix~\ref{app:baselines}.

\input{Table/table1}

\subsection{ARC-STAR as a Post-Hoc Corrector (RQ1)}
\label{sec:exp_main}

To answer \textbf{RQ1}, we compare \textsc{ARC-STAR} against 13 external baselines on the ten-cell suite under the matched-budget protocol of Sec.~\ref{sec:exp_setup}. Headline ratios are summarized in Table~\ref{tab:main_external}; we give the following observations (\textbf{Obs.}) over the ten-cell comparison:

\paragraph{Obs.~\ding{182} \textsc{ARC-STAR} reduces velocity rollout error by at least 36$\times$ on every cell.}
Across the ten cells, \textsc{ARC-STAR}'s ratios range from 0.0036 on NS-SL moderate to 0.0274 on KF moderate, and it stays below 0.03 on every cell. It is the best method on 7 of 10 cells. No baseline matches this consistency: ResNet exceeds $7\!\times\!10^{3}$ on NS-G extreme, PDE-Refiner stays above 0.3 on every cell under its recommended schedule, and PINN-Residual TTO exceeds the raw host on most cells.

\paragraph{Obs.~\ding{183} \textsc{ARC-STAR} dominates the frozen-host post-hoc family by one to two orders of magnitude.}
Within the family of methods sharing the frozen Poseidon host (Leray projection, PINN-Residual TTO, PINO-TTO-lite, DenseRes, and two parameter-efficient Poseidon-FT recipes), \textsc{ARC-STAR} improves over DenseRes by per-cell ratio reductions of 2.6$\times$ to 251$\times$, with the largest gaps on cells whose post-global residual is locally concentrated. Both Poseidon-FT recipes (Partial and LoRA, with full-parameter omitted as it requires per-cell learning-rate search beyond our matched-budget protocol) recover substantial host signal but stay above 0.030 on more than three cells; Leray helps only on NS-G; and PINO-TTO-lite improves the host only marginally. \textsc{ARC-STAR} is the strongest entry on every cell except KF moderate, a boundary case already flagged by the audit (Sec.~\ref{sec:exp_audit}).

\paragraph{Obs.~\ding{184} Boundary cells locate the gap rather than expose a failure mode.}
On three cells, NS-G $(m, x)$ and KF $(m)$, dense FNO or UNO trained from scratch reaches lower ratios than \textsc{ARC-STAR}. These are exactly the cells where the global stage already drives the residual ratio below 0.05, leaving a smooth, low-energy field that a dense Fourier operator can fit end-to-end. This pattern is consistent with limited local headroom rather than a routing failure, a reading made precise by the audit below. Data-efficiency curves on NS-PwC (15$\times$ reduction with as few as ten training trajectories, plateauing around one hundred) and rollout horizons up to $H{=}15$ are in Appendices~\ref{app:data-efficiency} and~\ref{app:extended-horizon}.

\subsection{ARC-STAR as a Stage-Wise Diagnostic (RQ2)}
\label{sec:exp_audit}
\input{Table/table2}
To answer \textbf{RQ2}, we apply the audit decomposition of Eq.~\eqref{eq:audit} to the ten-cell suite, summarized in Table~\ref{tab:stage_ablation}. We report medians of per-rollout quantities, so columns under ``Stage-wise recovery'' are not exact arithmetic transforms of the displayed median residual ratios; the protocol-level identity matches per-rollout values within 0.2$\%$ on every cell (Appendix~\ref{app:metric-stats}).

\paragraph{Obs.~\ding{185} The global stage carries most of the cleanup.} The global corrector alone removes 91$\%$ to 99$\%$ of raw host error on every cell, placing the post-global system at least an order of magnitude below the raw host across both viscosity regimes, even on the weakest cell. The post-global residual is therefore the natural starting point for reading the local stage's contribution.

\paragraph{Obs.~\ding{186} Local headroom predicts where the local stage helps.}
The local stage further removes 9.6$\%$ to 94.4$\%$ of the post-global residual; its largest gains land on cells whose post-global residual is most spatially concentrated in Fig.~\ref{fig:intro_motivation} (NS-PwC, NS-SL, and NS-Sines), and the same cells are those where \textsc{ARC-STAR}'s margin in Table~\ref{tab:main_external} is widest. On NS-G $(m, x)$ and KF $(m)$, post-global ratios are already small and $J_{\mathrm{loc}}$ is correspondingly low, exactly the boundary cells where dense Fourier baselines match or exceed \textsc{ARC-STAR} in Table~\ref{tab:main_external}. The audit therefore localizes each shortfall to a single stage rather than aggregating it into a suite-level statistic. A kinetic-energy spectrum check on NS-SL moderate (Appendix~\ref{app:ke-spectrum}) confirms full \textsc{ARC-STAR} matches GT to mean relative error $<$1$\%$ in the energy-bearing band, ruling out high-frequency artifacts from the local refiner.

\subsection{ARC-STAR under Compute Budgets (RQ3)}
\label{sec:exp_routing_qual}

To answer \textbf{RQ3}, we sweep deployment budgets $k/B \in \{0, 0.1, 0.2, 0.3, 0.4, 0.6, 0.8, 1.0\}$ and compare \textsc{ARC-STAR}'s label-free score against 9 external routing policies under shared host, global corrector, local refiner, and budget grid. Only block selection differs across runs.

\begin{figure}[!t]
\centering
\includegraphics[width=0.82\linewidth]{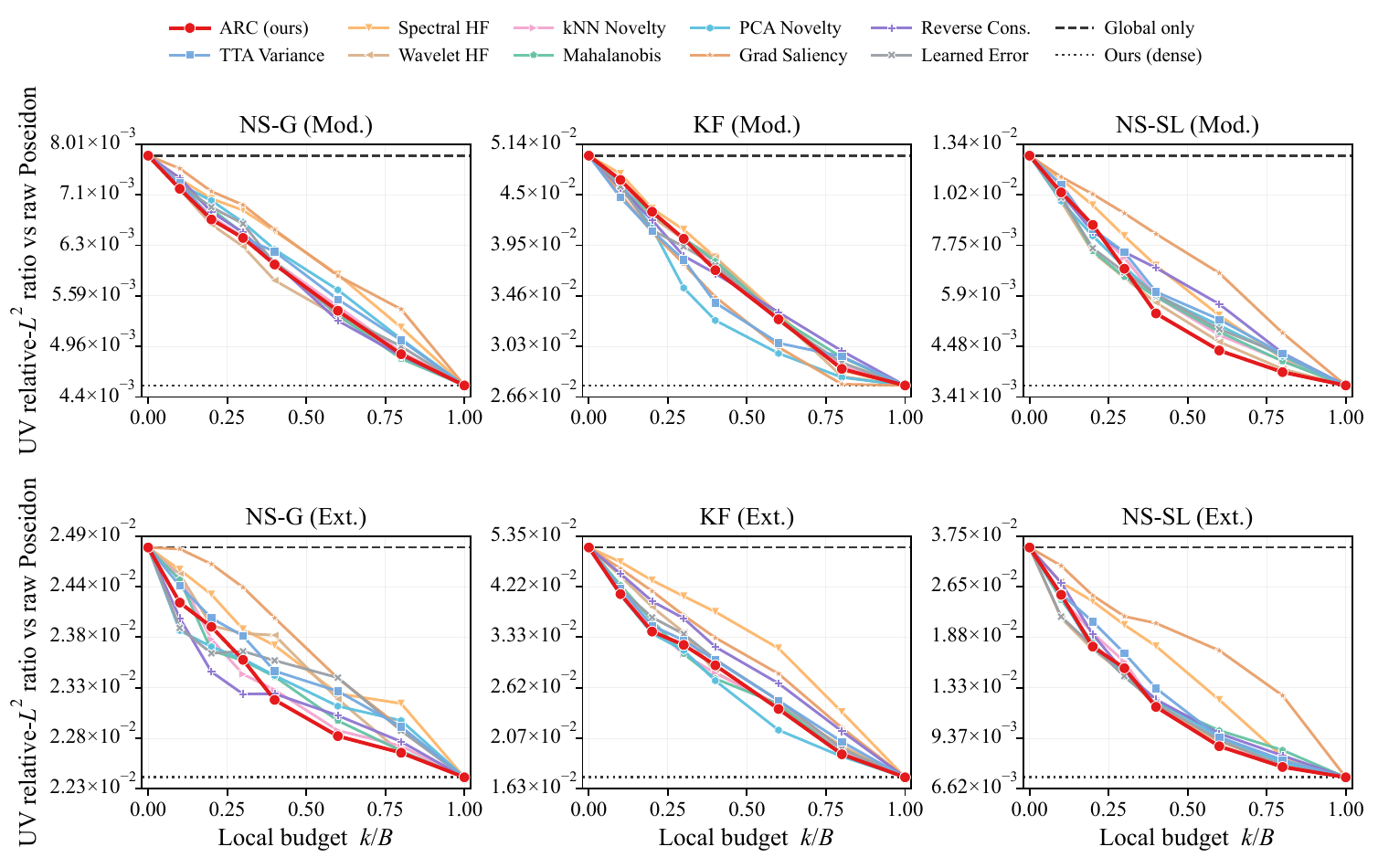}
\caption{\textbf{Routing frontier under shared compute.} \textsc{ARC-STAR} (red) versus 9 routing policies; only block selection differs. Dashed and dotted lines mark global-only ($k/B{=}0$) and dense \textsc{ARC-STAR} ($k/B{=}1$). Lower median 10-step UV-$L^2$ ratio is better.}
\label{fig:routing_frontier}
\end{figure}

\paragraph{Obs.~\ding{187} A single label-free score traces the lowest or near-lowest frontier on every cell.}
Figure~\ref{fig:routing_frontier} shows \textsc{ARC-STAR} below or matching every alternative at every budget on every cell, without per-cell tuning or inner-loop optimization. Saliency- and novelty-based alternatives swing across regimes, and the learned-error head leads on NS-G moderate but collapses to near-baseline on KF extreme. The full ten-cell frontier is in Figure~\ref{fig:routing_frontier_full} of Appendix~\ref{app:routing-frontier-full}.

\paragraph{Obs.~\ding{188} The audit doubles as a forward-only routing diagnostic.}
Cells with the highest $J_{\mathrm{loc}}$ in Table~\ref{tab:stage_ablation} are exactly those where \textsc{ARC-STAR} opens the widest margin over the global-only anchor in Fig.~\ref{fig:routing_frontier}; the cell with the lowest $J_{\mathrm{loc}}$ is where all nine external policies collapse into a 5$\%$ band. Routing skill is bounded by headroom rather than by the score, so a single training run at $k{=}B$ predicts how much routing-frontier headroom is recoverable on a given regime, before any deployment routing is invoked. A static-mask ablation (Appendix~\ref{app:static-mask}) confirms that the high mean per-step Jaccard (0.76) reflects underlying PDE structure rather than routing-rule redundancy.


\subsection{ARC-STAR across Frozen Hosts (RQ4)}
\label{sec:exp_cross_host}

\begin{figure}[!t]
\centering
\includegraphics[width=0.8\linewidth]{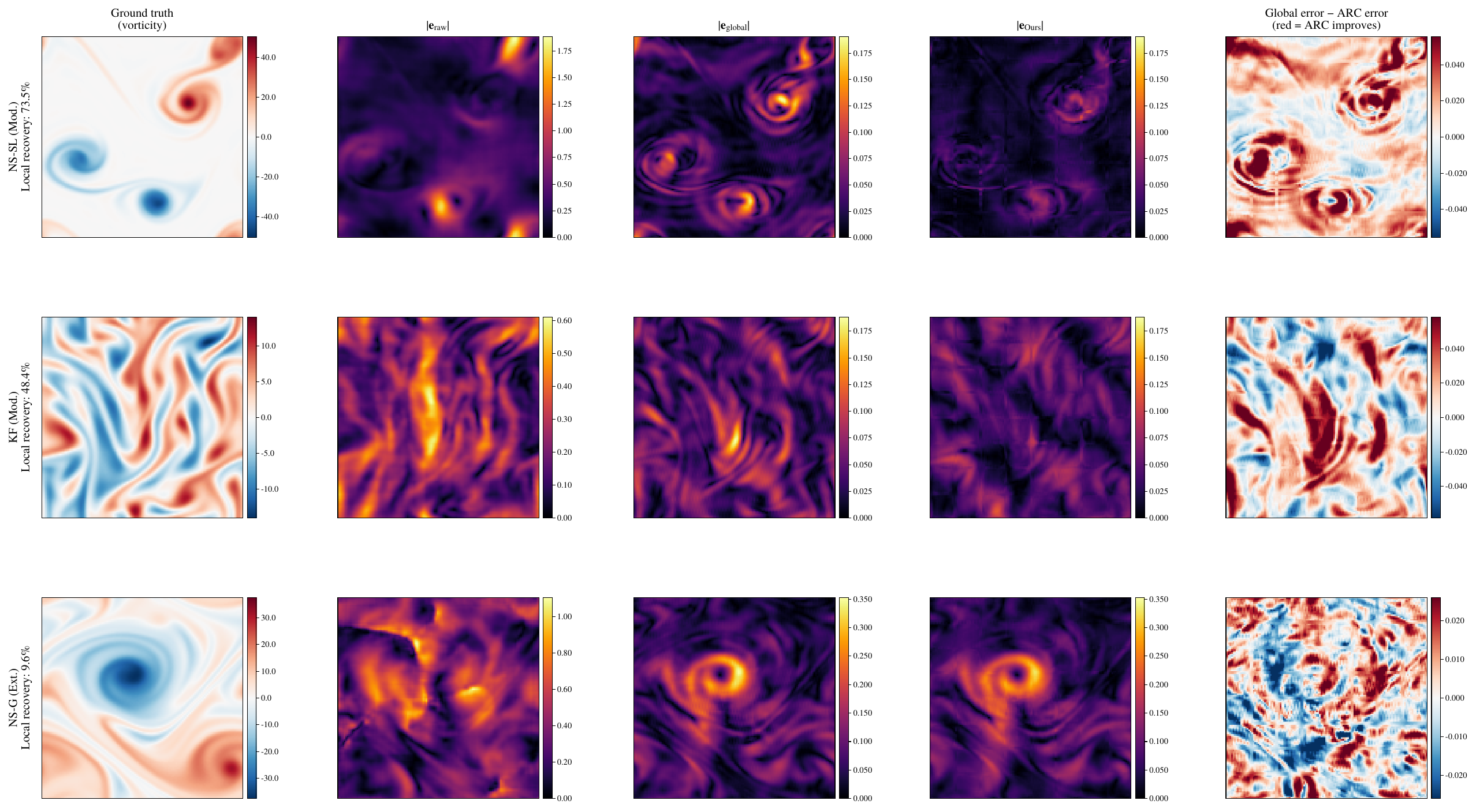}
\caption{\textbf{Qualitative rollout diagnosis.} Rows span high, medium, and low local recovery (NS-SL(m), KF(m), NS-G(x); $J_{\mathrm{loc}}\!=\!73.5\%, 48.4\%, 9.6\%$). The rightmost column shows local error reduction minus increase; strong reduction concentrates on structured vortex pockets, while NS-G(x) changes little, consistent with audit-bounded local headroom.}
\label{fig:qualitative}
\end{figure}

\paragraph{Obs.~\ding{189} The same recipe transfers to a structurally different frozen host.} Instantiating the same architecture template on a frozen DPOT-Ti~\citep{hao2024dpot} host with an AFNO backbone, with $G_\phi$ and $L_\theta$ trained from scratch under the identical recipe, reaches a 10-step UV relative-$L^2$ median ratio of 0.0139 against raw DPOT-Ti on NS-Gauss; cross-host kinetic-energy drift drops from $+$61$\%$ on the raw host to $-$5.3$\%$, against a ground-truth drift of $-$0.13$\%$. Figure~\ref{fig:qualitative} visualizes the same diagnosis in physical space: local refinement removes vortex-centered residuals on high-headroom cells and changes little on the boundary cell NS-G(x), consistent with the audit's headroom prediction. Cross-host details and routing-stability statistics are in Appendices~\ref{app:cross-host} and~\ref{app:routing-stability}. Cross-host portability is partial: the recipe transfers on global-stage-dominated cells but underperforms raw DPOT-Ti where the local refiner dominates (Appendix~\ref{app:cross-host-concentration}, host-specific overfitting).

\section{Conclusion}
\label{sec:conclusion}

In this paper, we introduce \textsc{ARC-STAR}, a deployment-time correction layer that wraps a frozen PDE foundation model with two trainable stages and a label-free spatial selector. \textsc{ARC-STAR} removes broad host bias with a global corrector and recovers locally concentrated residuals with a budget-aware blockwise refiner under a halo-read, center-write contract; the staged design lets practitioners attribute each unit of error reduction to either broad cleanup or local headroom, and lets a single trained module operate at any spatial budget without retraining. Across five incompressible-flow families with two viscosity regimes each, \textsc{ARC-STAR} consistently improves the frozen Poseidon host, transfers without recipe change to a structurally different DPOT-Ti host, and exposes the spatial-headroom regime in which routed correction is or is not the right tool. A potential limitation of \textsc{ARC-STAR} is its restriction to two-dimensional incompressible Navier-Stokes settings on periodic grids and to velocity-channel correction; extension to compressible, three-dimensional, or non-periodic regimes, longer rollout horizons, and host-modifying methods within a unified cross-paradigm benchmark suggest directions for future research on post-hoc correction at foundation scale.

\newpage

\bibliographystyle{plainnat}
\bibliography{reference}

\appendix
\input{appendix/A_inference}
\input{appendix/B_benchmarks}
\input{appendix/C_baselines}
\input{appendix/D_arcstar_details}

\input{appendix/E_latency}
\input{appendix/F_routing_stability}

\input{appendix/G_divergence}
\input{appendix/H_cross_host}
\input{appendix/J_static_mask}
\input{appendix/K_pderefiner_sensitivity}


\end{document}

%% file: Table/table1.tex
\begin{table}[t]
\centering
\caption{\textbf{10-step UV relative-$L^2$ ratios vs.\ raw Poseidon on five incompressible-flow benchmarks} (lower is better). \hlfirst{Best} and \hlsecond{runner-up} per cell. Method blocks (separated by horizontal rules, top to bottom): \textit{raw host}; \textit{from-scratch dense operators} (no Poseidon prior); \textit{iterative refinement}; \textit{host-modifying baseline}; \textit{frozen-Poseidon post-hoc family} (head-to-head with ours); and \textsc{ARC-STAR}. Full budget sensitivity in Appendix~\ref{app:budget-sweep}. The (m) and (x) sub-columns refer to the moderate and extreme viscosity regimes of each benchmark.}
\label{tab:main_external}
{\small
\setlength{\tabcolsep}{4pt}
\renewcommand{\arraystretch}{1.1}
\resizebox{\linewidth}{!}{
\begin{tabular}{l|cc|cc|cc|cc|cc}
\hline
 & \multicolumn{2}{c|}{\textbf{NS-SL}} & \multicolumn{2}{c|}{\textbf{KF}} & \multicolumn{2}{c|}{\textbf{NS-PwC}} & \multicolumn{2}{c|}{\textbf{NS-G}} & \multicolumn{2}{c}{\textbf{NS-Sines}} \\
 & (m) & (x) & (m) & (x) & (m) & (x) & (m) & (x) & (m) & (x) \\
\hline
Poseidon (raw)~\citep{herde2024poseidon} & 1.000 & 1.000 & 1.000 & 1.000 & 1.000 & 1.000 & 1.000 & 1.000 & 1.000 & 1.000 \\
\hline
FNO~\citep{li2021fourier} & 0.0248 & 0.0355 & 0.0103 & 0.0459 & 0.101 & 0.185 & \hlfirst{0.0033} & \hlfirst{0.0049} & 0.325 & 0.527 \\
TFNO~\citep{kossaifi2024tfno} & 0.0308 & 0.0378 & \hlsecond{0.0101} & \hlsecond{0.0180} & 1.113 & 1.395 & 0.0210 & \hlsecond{0.0163} & 0.489 & 0.668 \\
UNO~\citep{rahman2023uno} & \hlsecond{0.0148} & \hlsecond{0.0180} & \hlfirst{0.0084} & \hlfirst{0.0172} & 0.0109 & \hlsecond{0.0184} & 0.0235 & 0.0232 & 0.116 & 0.188 \\
U-Net~\citep{ronneberger2015unet} & 0.270 & 0.298 & 0.137 & 0.335 & 1.155 & 1.441 & 2.821 & 3.145 & 0.755 & 0.997 \\
ResNet~\citep{he2016resnet} & 2.154 & 2.527 & 0.471 & 0.947 & 1.125 & 1.395 & 5.95$\times 10^{3}$ & 7.30$\times 10^{3}$ & 1.208 & 1.574 \\
PDEArena~\citep{gupta2023pdearena} & 4.374 & 3.817 & 8.113 & 7.106 & 7.199 & 6.252 & 5.119 & 3.870 & 2.660 & 2.868 \\
\hline
PDE-Refiner~\citep{lippe2023pderefiner} & 0.313 & 0.314 & 1.099 & 1.224 & 1.265 & 1.508 & 0.474 & 0.401 & 0.479 & 0.614 \\
\hline
Poseidon-FT (Partial)~\citep{herde2024poseidon} & 0.0417 & 0.0595 & 0.0391 & 0.0352 & 0.0405 & 0.0709 & 0.0644 & 0.0664 & 0.1181 & 0.1455 \\
Poseidon-FT (LoRA $r{=}8$)~\citep{hu2022lora} & 0.0944 & 0.1316 & 0.0135 & 0.0274 & \hlsecond{0.0090} & 0.0432 & 0.0443 & 0.0740 & 0.1418 & \hlsecond{0.0827} \\ 
\hline
Leray projection~\citep{chorin1968numerical} & 1.035 & 1.068 & 1.005 & 1.017 & 1.011 & 1.013 & 0.689 & 0.433 & 0.990 & 0.982 \\
PINN-Residual TTO~\citep{raissi2019physics} & 2.744 & 2.559 & 2.436 & 2.705 & 2.532 & 2.272 & 2.978 & 0.902 & 2.740 & 2.967 \\
PINO-TTO-lite~\citep{li2021pino} & 0.665 & 0.733 & 0.601 & 0.626 & 0.583 & 0.653 & 0.955 & 0.867 & 0.865 & 0.970 \\
DenseRes & 0.644 & 0.816 & 0.0689 & 0.180 & 0.0395 & 0.0768 & 1.105 & 0.775 & \hlsecond{0.0757} & 0.0865 \\
\hline
\textbf{\textsc{ARC-STAR} (Ours)} & \hlfirst{0.0036} & \hlfirst{0.0072} & 0.0274 & \hlfirst{0.0172} & \hlfirst{0.0047} & \hlfirst{0.0056} & \hlsecond{0.0045} & 0.0224 & \hlfirst{0.0162} & \hlfirst{0.0213} \\
\hline
\end{tabular}
}}
\end{table}

%% file: Table/table2.tex
\begin{table}[t]
\centering
\caption{Stage-wise audit decomposition of \textbf{ARC-STAR} across the moderate (Mod.) and extreme (Ext.) regimes of each benchmark, matching the (m) / (x) split used in Table~\ref{tab:main_external}. Residual ratio columns report $L/L^{\mathrm{raw}}$ at the raw, post-global, and post-ARC-STAR stages; Stage-wise recovery columns report per-stage cleanup against each stage's own input baseline. }
\label{tab:stage_ablation}
\footnotesize
\setlength{\tabcolsep}{4pt}
\renewcommand{\arraystretch}{1.05}
\begin{tabular*}{\linewidth}{@{\extracolsep{\fill}}llccc|cc@{}}
\toprule
& & \multicolumn{3}{c|}{Residual ratio (vs.\ Raw)} & \multicolumn{2}{c}{Stage-wise recovery (\%)} \\
\cmidrule(lr){3-5} \cmidrule(lr){6-7}
Benchmark & Regime & Raw & Global only & \textbf{ARC-STAR} & Recovery from Raw & Recovery from Post-Global \\
\midrule
\multirow{2}{*}{NS-SL}
  & Mod. & 1.000 & 0.0126 & 0.0036 & 98.7 & 73.5 \\
  & Ext. & 1.000 & 0.0347 & 0.0072 & 96.5 & 74.4 \\
\midrule
\multirow{2}{*}{KF}
  & Mod. & 1.000 & 0.0499 & 0.0274 & 95.0 & 48.4 \\
  & Ext. & 1.000 & 0.0507 & 0.0172 & 94.9 & 62.8 \\
\midrule
\multirow{2}{*}{NS-PwC}
  & Mod. & 1.000 & 0.0714 & 0.0047 & 92.9 & 94.4 \\
  & Ext. & 1.000 & 0.0497 & 0.0056 & 95.0 & 89.0 \\
\midrule
\multirow{2}{*}{NS-G}
  & Mod. & 1.000 & 0.0078 & 0.0045 & 99.2 & 43.9 \\
  & Ext. & 1.000 & 0.0248 & 0.0224 & 97.5 & 9.6 \\
\midrule
\multirow{2}{*}{NS-Sines}
  & Mod. & 1.000 & 0.0742 & 0.0162 & 92.6 & 76.5 \\
  & Ext. & 1.000 & 0.0886 & 0.0213 & 91.1 & 76.0 \\
\bottomrule
\end{tabular*}
\end{table}

%% file: appendix/A_inference.tex
\section{Inference Procedure and Notation}
\label{app:inference}

\subsection{Notation}
\label{app:notation}

\begin{table}[h]
\centering
\caption{Notation used throughout the paper.}
\label{tab:notation}
\small
\begin{tabular}{ll}
\toprule
Symbol & Meaning \\
\midrule
$x_t$, $x^*_{t+1}$ & State at time $t$; ground-truth state at $t{+}1$ \\
$H, \hat x_{t+1}$ & Frozen pretrained PDE host; raw host forecast \\
$G_\phi$, $L_\theta$ & Global corrector (trainable $\phi$); local refiner (trainable $\theta$) \\
$x^g_{t+1}$, $x^{\mathrm{hyb}}_{t+1}$ & Post-global field (Eq.~\ref{eq:global}); hybrid output (Eq.~\ref{eq:local}) \\
$\Omega_b$, $\Omega_b^{+h}$ & Center block $b$ ($b{\times}b$); halo-extended block ($(b{+}2h){\times}(b{+}2h)$) \\
$\mathcal{C}$ & Center-crop operator: $(b{+}2h){\times}(b{+}2h) \to b{\times}b$ \\
$\mathcal{S}_t \subseteq \{1,\ldots,B\}$ & Active block set at step $t$ \\
$r_t(p)$, $s_b(t)$ & Pixel-level risk; block-aggregated score (Eq.~\ref{eq:risk},~\ref{eq:score}) \\
$\lambda_{\mathrm{KE}}$ & Kinetic-energy weighting in the risk score (fixed to $0.05$) \\
$T$, $d_v$, $m$, $\alpha$ & TFNO-based: \#Fourier layers, lifting dim, spectral modes, residual scale \\
$b$, $h$ & ConvNeXt: block edge length, halo width \\
$k$, $B$ & Top-$k$ size; total \#blocks per field ($B{=}(128/b)^2$) \\
$L^{\mathrm{raw}}, L^{\mathrm{glob}}, L^{\mathrm{hyb}}$ & Rollout errors of host / global-only / hybrid \\
$A$, $J_{\mathrm{loc}}$ & Audit terms (Eq.~\ref{eq:audit_terms}) \\
\bottomrule
\end{tabular}
\end{table}

\subsection{Single-Step Inference}
\label{app:pseudocode}

\begin{algorithm}[H]
\caption{\textsc{ARC-STAR} single rollout step}
\begin{algorithmic}[1]
\State \textbf{Input:} state $x_t$; frozen $H, G_\phi, L_\theta$; budget $k \in \{1,\ldots,B\}$
\State $\hat x_{t+1} \leftarrow H(x_t)$ \Comment{frozen host forward}
\State $x^{g}_{t+1} \leftarrow \hat x_{t+1} + \Pi_{uv}\!\left[G_\phi(x_t, \hat x_{t+1})\right]$ \Comment{Eq.~\eqref{eq:global}}
\State $r_t \leftarrow |\Delta u_t| + |\Delta v_t| + \lambda_{\mathrm{KE}}\!\left(|\partial_x K^{g}_{t+1}| + |\partial_y K^{g}_{t+1}|\right)$ \Comment{Eq.~\eqref{eq:risk}}
\State $\{s_b(t)\}_{b=1}^{B} \leftarrow \texttt{block\_mean\_map}(r_t, b)$ \Comment{Eq.~\eqref{eq:score}}
\If{$k = B$} \Comment{dense mode}
    \State $\mathcal{S}_t \leftarrow \{1,\ldots,B\}$
\Else \Comment{budgeted mode}
    \State $\mathcal{S}_t \leftarrow \texttt{TopK}(\{s_b(t)\}, k)$
\EndIf
\For{$b \in \mathcal{S}_t$}
    \State $W_b \leftarrow \texttt{halo\_extract}(x_t, x^{g}_{t+1}; \Omega_b^{+h})$
    \State $\Pi_{uv}\!\left(x^{\mathrm{hyb}}_{t+1}\right)\!\restriction_{\Omega_b} \leftarrow \Pi_{uv}\!\left(x^{g}_{t+1}\right)\!\restriction_{\Omega_b} + w_{\mathrm{H}} \odot \mathcal{C}\!\left[L_\theta(W_b)\right]$ \Comment{Eq.~\eqref{eq:local}}
\EndFor
\State \textbf{for} $b \notin \mathcal{S}_t$: $x^{\mathrm{hyb}}_{t+1}\!\restriction_{\Omega_b} \leftarrow x^{g}_{t+1}\!\restriction_{\Omega_b}$ \Comment{pass-through}
\State \Return $x^{\mathrm{hyb}}_{t+1}$
\end{algorithmic}
\end{algorithm}

The center-crop $\mathcal{C}$ enforces the halo-read, center-write contract: $L_\theta$ reads context from $\Omega_b^{+h}$ but only writes the centered $b{\times}b$ region, so unselected blocks are bit-exactly equal to $x^g_{t+1}$ (no boundary blending artifacts on the unselected region).

%% file: appendix/B_benchmarks.tex
\section{Benchmark Details}
\label{app:benchmarks}

\subsection{Selection Rationale}

We evaluate on five 2D incompressible-flow families from the public Poseidon benchmark suite~\citep{herde2024poseidon}, selected to span representative incompressible-flow regimes:

\begin{itemize}
    \item NS-G (Gaussian initial conditions): tests broadband-spectrum smooth fields.
    \item KF (Kolmogorov forcing): probes mixed-scale stationary turbulence with persistent forcing.
    \item NS-SL (shear layer): injects a sharp tangential discontinuity that develops Kelvin-Helmholtz instabilities.
    \item NS-PwC (piecewise-constant initial vorticity): creates compactly supported errors at vortex boundaries.
    \item NS-Sines (sinusoidal initial conditions): exposes wavenumber leakage between principal Fourier modes.
\end{itemize}

These five families cover smooth, forced-turbulent, sharp-discontinuity, piecewise-constant, and band-limited flows respectively, providing a representative slice of 2D incompressible Navier-Stokes dynamics. Each family is evaluated under a moderate (m) and an extreme (x) regime along the viscosity axis, giving 10 benchmark--regime cells in total.

\subsection{Numerical Setup}

All benchmarks share the same numerical setup: a periodic $2\pi \times 2\pi$ square domain discretized on a $128 \times 128$ uniform grid, integrated with a fixed time step $\Delta t = 0.05$. The moderate and extreme regimes share initial-condition distributions but differ by one order of magnitude in viscosity.

\begin{table}[h]
\centering
\caption{Numerical setup shared across all five benchmarks. The moderate and extreme regimes differ only in viscosity (and therefore characteristic Reynolds number); all other quantities are identical.}
\label{tab:benchmarks-numerics}
\small
\begin{tabular}{lcc}
\toprule
Quantity & Moderate (m) & Extreme (x) \\
\midrule
Viscosity $\nu$ & $1.0 \times 10^{-3}$ & $1.0 \times 10^{-4}$ \\
Characteristic Reynolds $\mathrm{Re}$ & $\sim\!10^{3}$ & $\sim\!10^{4}$ \\
Domain & $[0, 2\pi]^2$ (periodic) & $[0, 2\pi]^2$ (periodic) \\
Grid resolution & $128 \times 128$ & $128 \times 128$ \\
Time step $\Delta t$ & $0.05$ & $0.05$ \\
Rollout horizon (evaluation) & 10 steps & 10 steps \\
\bottomrule
\end{tabular}
\end{table}

\subsection{Initial-Condition Distributions}

\begin{table}[h]
\centering
\caption{Per-benchmark forcing and initial-condition distribution. Each family ships with roughly $1500$--$2000$ pre-simulated trajectories from the Poseidon benchmark suite, from which the training and test splits below are drawn.}
\label{tab:benchmarks-fams}
\small
\begin{tabular}{lll}
\toprule
Family & Forcing & Initial-condition distribution \\
\midrule
NS-G     & none               & isotropic Gaussian random fields \\
KF       & Kolmogorov forcing & equilibrium-drawn from forced runs \\
NS-SL    & none               & shear layer with two opposing strips \\
NS-PwC   & none               & piecewise-constant vorticity blobs \\
NS-Sines & none               & low-mode sinusoidal superpositions \\
\bottomrule
\end{tabular}
\end{table}

\subsection{Train and Test Splits}

For each benchmark, we draw $N_{\mathrm{train}} = 200$ trajectories for training the global corrector and the local refiner (\texttt{-train\_n 200}). For evaluation we use a four-trajectory subset drawn from Poseidon's official held-out test split (which contains $\sim$240 test trajectories per cell at the released split sizes 19,640/120/240 for NS-G/KF/NS-PwC/NS-Sines and 39,640/120/240 for NS-SL). The four indices per cell are listed verbatim in Table~\ref{tab:benchmarks-splits}; subsampling here is computational rather than statistical, chosen to keep the cross-baseline matrix tractable at $n{=}8$ paired runs per cell, and the subset is fixed once and held constant across every method, regime, and ablation. Train and test splits are disjoint within each benchmark. Within the 200 training trajectories, the last 20 are reserved as a validation set for early-stopping (patience 30 epochs); validation trajectories never overlap with the four canonical test trajectories listed in Table~\ref{tab:benchmarks-splits}, so early-stopping decisions cannot leak test-time signal.

\begin{table}[h]
\centering
\caption{Held-out test-trajectory indices for each benchmark--regime cell. Each row lists the four trajectory indices on which evaluation is performed; pairing each trajectory with $t_0 \in \{5, 10\}$ yields $n = 8$ runs per cell, used for the median-of-ratios statistic (Appendix~\ref{app:metric-stats}). Indices are taken verbatim from the canonical split files emitted during data preparation (\texttt{summary.json::split::\{moderate,extreme\}\_test\_indices}).}
\label{tab:benchmarks-splits}
\small
\begin{tabular}{lll}
\toprule
Benchmark & Moderate (m) test indices & Extreme (x) test indices \\
\midrule
NS-G     & 596, 817, 341, 438    & 791, 451, 846, 580 \\
KF       & 1632, 1665, 756, 1484 & 664, 532, 1591, 290 \\
NS-SL    & 817, 789, 1322, 1458  & 648, 129, 201, 1204 \\
NS-PwC   & 707, 963, 639, 121    & 245, 664, 532, 290 \\
NS-Sines & 1009, 529, 724, 666   & 267, 712, 390, 978 \\
\bottomrule
\end{tabular}
\end{table}

\paragraph{Cross-benchmark index reuse.} A small number of trajectory indices appear in multiple cells (e.g.\ index $290$ in both KF~(x) and NS-PwC~(x)). These are not leaks: each benchmark family draws from its own trajectory pool generated under that family's IC distribution and forcing, so the same integer index identifies \emph{different} physical trajectories across families.

\subsection{Sampled Starting Times}

\begin{itemize}
    \item \textbf{Training-time sampling:} during patch extraction and AR fine-tuning, starting times are drawn from $\{3, 5, 7, 10, 13\}$ (\texttt{--train\_t0s}). This spans a range of rollout depths so the refiner sees both early-stage fields (still close to the IC) and later-stage fields (where host error has accumulated).
    \item \textbf{Evaluation:} every test trajectory is evaluated at $t_0 \in \{5, 10\}$. With four test trajectories per cell, this gives $n = 8$ runs.
\end{itemize}

\subsection{Pre-processing}

\paragraph{Input channels.} The state vector has 4 channels $(u, v, \cdot, \cdot)$, where the first two are the velocity components used by the metric and the last two carry passive scalar or pressure information from the Poseidon convention. Inside the TFNO-based forward pass, two sinusoidal grid coordinates are concatenated to the input (\texttt{cstar\_global\_tfno\_corrector.py:31--34}), producing 10 input channels into the lifting projection $P$.

\paragraph{Periodic boundary handling.} All halo extractions and spectral convolutions assume periodic boundaries consistent with the $2\pi$ domain. Halo padding for $\Omega_b^{+h}$ uses circular boundary conditions (\texttt{cstar\_pde\_minimal.py:130}); the FFT inside each Fourier layer assumes a periodic field by construction.

\paragraph{Normalization.} The reported metric is the relative $L^2$ error on $(u, v)$ averaged over the spatial domain, normalized per run by the raw Poseidon rollout on the same trajectory (Sec.~\ref{sec:exp_setup}). No additional channel-wise normalization is applied to the state itself; per-run normalization controls for trajectory-level difficulty variation without imposing dataset-level statistics.

\subsection{Run Definition}

A single \emph{run} is a 10-step autoregressive rollout starting from one held-out trajectory at one starting time $t_0$. With four trajectories per cell and $t_0 \in \{5, 10\}$, each cell contributes $n = 8$ runs. Reported numbers throughout the paper are medians of per-run quantities over these eight runs (Appendix~\ref{app:metric-stats}); paired bootstrap 95\% confidence intervals on each cell are reported in Table~\ref{tab:bootstrap-ci}.

\subsection{Absolute raw rollout error per cell}
\label{app:absolute-raw}

All normalized ratios in Table~\ref{tab:stage_ablation} share a common per-cell denominator $L^{\text{raw}}$, the absolute RMSE of the frozen Poseidon rollout on that cell. Table~\ref{tab:raw_rmse} reports these absolute denominators so that any normalized ratio in the main paper can be converted back to physical units. We note one observation that informs the interpretation of NS-SL: the moderate and extreme regimes of NS-SL share an almost identical raw RMSE ($3.14 \times 10^{-2}$ in both), even though their post-global residuals differ by roughly a factor of three. Severity of distribution shift therefore manifests primarily in the post-global residual rather than in the raw rollout itself for this benchmark.

\begin{table}[h]
\centering
\caption{Absolute raw rollout relative $L^2$ error per (benchmark, regime) cell, used as the denominator $L^{\text{raw}}$ throughout the paper. Per-run medians over $n=8$ runs (4 trajectories $\times$ 2 initial times).}
\label{tab:raw_rmse}
\footnotesize
\begin{tabular*}{\linewidth}{@{\extracolsep{\fill}}lcc@{}}
\toprule
Benchmark & Mod. regime & Ext. regime \\
\midrule
NS-SL    & $3.14 \times 10^{-2}$ & $3.14 \times 10^{-2}$ \\
KF       & $1.34 \times 10^{-2}$ & $1.64 \times 10^{-2}$ \\
NS-PwC   & $1.28 \times 10^{-2}$ & $1.59 \times 10^{-2}$ \\
NS-G     & $4.83 \times 10^{-2}$ & $1.60 \times 10^{-1}$ \\
NS-Sines & $3.77 \times 10^{-2}$ & $4.46 \times 10^{-2}$ \\
\bottomrule
\end{tabular*}
\end{table}

%% file: appendix/C_baselines.tex
\section{Baseline Implementation Details}
\label{app:baselines}

\paragraph{Two groups of comparisons.}
The methods in Table~\ref{tab:main_external} fall into two categories. \emph{External baselines} (FNO, TFNO, UNO, U-Net, ResNet, PDEArena, PDE-Refiner, Leray projection) are independent dense or projection-based methods evaluated using their published recipes. \emph{Implementation-controlled references} (Poseidon-FT, DenseRes, PINO-TTO-lite, PINN-Residual TTO, and the Learned Error Head used in Fig.~\ref{fig:routing_frontier}) are matched-budget controls we construct to test specific hypotheses against the same Poseidon host, the same $200$ training trajectories, and the same evaluation protocol: DenseRes tests whether a single-stage residual correction is sufficient (vs.\ two-stage); PINO-TTO-lite and PINN-Residual TTO test whether physics-informed test-time optimization can substitute for a learned local refiner; the Learned Error Head tests whether a label-supervised score can outperform our label-free $s_b(t)$. We do not claim these references represent the strongest published instantiations of their respective ideas; they are isolated to make any performance gap attributable to the hypothesis under test, not to architectural or training-budget asymmetries.

\subsection{External Baselines (Table~\ref{tab:main_external})}
\label{app:external-baselines}

We group the twelve external baselines into four families: dense neural operators, iterative refinement, post-hoc correction, and test-time correction.

\subsubsection{Dense Neural Operators}

\paragraph{FNO~\citep{li2021fourier}.}
Standard Fourier Neural Operator with spectral convolutions in Fourier space. We follow the original NS-2D configuration: 4 Fourier layers, 12 spectral modes per axis, 32 hidden channels, GELU activation, with a $1{\times}1$ input lifting and $1{\times}1$ output projection. Trained from scratch on each cell's 200 training trajectories with Adam (learning rate $10^{-3}$, batch size 8, cosine schedule, 200 epochs, early stopping patience 30), under the same 5-step AR closed-loop loss as ARC-STAR's global corrector.

\paragraph{TFNO~\citep{kossaifi2024tfno}.}
Tucker-factorized FNO with the same depth and width as FNO above (4 layers, 32 hidden channels, 12 modes), and Tucker rank 8 on the spectral weight tensor. Training protocol identical to FNO.

\paragraph{UNO~\citep{rahman2023uno}.}
Hierarchical U-shaped FNO with three resolution scales connected by $2\times$ down/up-sampling. Each scale uses 32 hidden channels and 12 spectral modes. Skip connections at matching resolutions. Training protocol identical to FNO.

\paragraph{U-Net~\citep{ronneberger2015unet}.}
Three-level encoder--decoder convolutional architecture: 64 base channels, $3{\times}3$ convolutions, GroupNorm, GELU activation, $2\times$ down/up-sampling, skip connections at matching resolutions. Training protocol identical to FNO.

\paragraph{ResNet~\citep{he2016resnet}.}
8 residual blocks with 64 channels each (two $3{\times}3$ convolutions plus identity skip per block, GroupNorm, GELU). A $3{\times}3$ input projection lifts the 4-channel state to 64 channels and a $1{\times}1$ output projection produces the 2-channel prediction. Training protocol identical to FNO.

\paragraph{PDE-Arena~\citep{gupta2023pdearena}.}
The U-Net variant from the PDE-Arena release with the default 4-level architecture and 64 base channels. Training protocol identical to FNO.

\subsubsection{Iterative Refinement}

\paragraph{PDE-Refiner~\citep{lippe2023pderefiner}.}
Diffusion-style iterative refinement. Table~\ref{tab:main_external} reports the Recommended-8 + EMA configuration recommended by \citet{lippe2023pderefiner} for incompressible-flow benchmarks (8 refinement steps, EMA decay 0.995, $\sigma_{\min}=2\times 10^{-3}$, $\sigma_{\max}=0.5$); the Default-4 configuration (4 steps, no EMA, $\sigma_{\max}=0.2$) is reported alongside in Appendix~\ref{app:pderefiner-sensitivity} for completeness. The base denoising network matches the FNO architecture above (4 Fourier layers, 32 hidden channels). Training and rollout protocol are identical to FNO.

\subsubsection{Post-Hoc Correction}

\paragraph{Leray projection~\citep{chorin1968numerical}.}
Training-free post-processing operator that projects the raw host forecast onto the divergence-free subspace via a Helmholtz--Hodge decomposition (FFT-based pressure solve under periodic boundaries). No training; applied at every rollout step on top of the raw host output.

\paragraph{Poseidon-FT (full-param).}
Full fine-tuning of the Poseidon host on each cell's 200 training trajectories. We use AdamW (learning rate $10^{-4}$, weight decay $10^{-4}$), batch size 4, 100 epochs with early stopping (patience 20), under the same 5-step AR closed-loop loss as ARC-STAR.

\paragraph{Poseidon-FT (Partial).}
Parameter-efficient fine-tuning following the recommended recipe of \citet{herde2024poseidon}: we freeze the Swin-v2 attention blocks and unfreeze only the time-conditioning \textsc{ConditionalLayerNorm} weights and biases (\texttt{layernorm\_before/after}) across all encoder/decoder blocks together with the input \texttt{embeddings.norm}, the last decoder stage \texttt{decoder.layers.3}, and the output head \texttt{patch\_recovery}, yielding $174{,}176$ trainable parameters out of $20.77\mathrm{M}$ total ($0.84\%$). We use AdamW with learning rate $10^{-4}$, weight decay $10^{-4}$, cosine schedule, batch size 4, patience-15 early stopping, and a maximum of $80$ epochs; median stopping epoch was $27.5$. Training data and the AR closed-loop loss are identical to the full-parameter setting. Both Poseidon-FT variants update the host parameters and use the fine-tuned host as the rollout backbone at evaluation, distinguishing them from all post-hoc comparisons.

\paragraph{Poseidon-FT (LoRA $r{=}8$).}
LoRA fine-tuning~\citep{hu2022lora} applied to the same time-conditioning \textsc{ConditionalLayerNorm} parameters as Partial-FT, with rank $r{=}8$. This yields ${\sim}273\mathrm{K}$ trainable parameters out of the same $20.77\mathrm{M}$ total ($1.31\%$). Training data, optimizer hyperparameters, schedule, and patience are identical to the Partial-FT setting; only the parameterization of the trainable weights differs.

\paragraph{DenseRes.}
An implementation-controlled, lightweight post-hoc reference. Given $(x_t, \hat x_{t+1})$, DenseRes predicts a single-stage residual $r_{\mathrm{INC}}(x_t, \hat x_{t+1})$ that is added back to the host forecast (no global/local decomposition). We instantiate it as a 4-layer residual CNN with GroupNorm: depth $=4$ ResBlocks, base channels $=64$, total parameter count $=301{,}378$ ($0.30$M). Each ResBlock consists of two $3{\times}3$ convolutions sandwiched by GroupNorm and GELU activations, with an identity skip. A $3{\times}3$ input projection ($8 \to 64$ channels, taking $[x_t, \hat x_{t+1}]$ as input) precedes the residual stack; a GroupNorm--GELU--$1{\times}1$ output projection produces the 2-channel residual. Training: AdamW with learning rate $10^{-3}$, batch size 2, 60 epochs with early stopping (patience 15), on the same 200 canonical training trajectories as ARC-STAR's global corrector. Total compute is below $0.5$ GPU-hours per benchmark on a single A100~40GB, well within ARC-STAR's combined Stage-1 plus Stage-2 budget.

\subsubsection{Test-Time Correction}

\textbf{Newton--Krylov PINN residual corrector (NK-PINN).} A test-time PINN-style residual corrector adapted from the PINN formulation of Raissi et al.~\cite{raissi2019physics}. At each rollout step, given the host forecast $\hat{x}_{t+1}$, we minimize a PDE-residual loss for incompressible Navier--Stokes (continuity plus momentum) over a small correction $\delta$ via 50 Newton--Krylov inner iterations starting from $\delta = 0$. Inner-loop linear systems are solved via 20 GMRES iterations with Jacobi preconditioning. \emph{Note:} this is a compute-matched approximation of physics-residual test-time correction; we do \emph{not} evaluate the cached-Jacobian PhysicsCorrect variant of Huang and Perdikaris~\cite{huang2026physicscorrect}, whose offline pseudoinverse warm-up requires per-cell Jacobian factorization outside our matched-budget protocol. Comparison against the cached-Jacobian variant is left to future work.

\paragraph{Conceptual contrast with PhysicsCorrect.} PhysicsCorrect~\citep{huang2026physicscorrect} is a training-free test-time PDE-residual correction that solves a per-step linearization of the governing equations against a cached Jacobian. Both methods share the frozen-host commitment but differ on three axes: (i)~training status: PhysicsCorrect requires no learned correction stage, while ARC-STAR trains a global corrector and a local refiner; (ii)~deployment-time compute: PhysicsCorrect performs a per-step pseudoinverse application, while ARC-STAR uses a forward pass through pre-trained networks; and (iii)~per-cell setup: the cached-Jacobian variant requires per-cell offline factorization, while ARC-STAR's $G_\phi$ and $L_\theta$ are trained per cell but reused across budgets without retraining. A direct empirical comparison would require either an uncached PhysicsCorrect run (50+ GPU-hours per cell) or per-cell Jacobian factorization (outside our matched-budget protocol); we leave this comparison to follow-up work. We acknowledge that this matched-budget protocol excludes PhysicsCorrect's cached-Jacobian variant, which reports up to $100\times$ error reduction at sub-$5\%$ inference overhead on different benchmarks. A direct ARC-STAR vs PhysicsCorrect (cached) comparison on a shared benchmark and rollout protocol is left to future work; the two methods are not mutually exclusive—ARC-STAR's halo-read center-write contract could in principle wrap a cached-Jacobian global stage.

\paragraph{PINO-TTO-lite~\citep{li2021pino}.}
Test-time optimization with a physics-informed loss following PINO. Starting from the frozen host forecast $\hat x_{t+1}$, we perform 50 inner Adam steps (learning rate $10^{-3}$) on a parametric correction $\delta_\psi(x_t, \hat x_{t+1})$ realized as a small 2-layer convolutional network ($\sim 20$K parameters) re-initialized at every rollout step. The loss combines an MSE-to-host term and a PDE-residual term with equal weight. No training; applied test-time.

\subsection{External Routing Policies (Fig.~\ref{fig:routing_frontier})}
\label{app:routing-policies}

The nine external policies share the same frozen host, global corrector, local refiner, rollout protocol, and budget grid as ARC's \texttt{innovation\_keg} score; only the per-block scoring rule differs. All scores are aggregated to per-block scalars by mean-pooling within $\Omega_b$ (\texttt{block\_mean\_map}, \texttt{cstar\_pde\_minimal.py:121--123}), and the top-$k$ blocks are selected under the prescribed budget $k/B$.

\subsubsection{Uncertainty}

\paragraph{TTA-variance~\citep{gal2016dropout,lakshminarayanan2017simple,ayhan2018test}.}
Per-pixel variance under 8 test-time augmentations: random horizontal/vertical flips and $\{0^\circ, 90^\circ, 180^\circ, 270^\circ\}$ rotations applied to the input. Each augmented input is passed through the frozen $G_\phi$, predictions are inverse-mapped to the original frame, and the per-pixel variance across the 8 augmented predictions is taken as the score.

\subsubsection{Signal-Domain Detail}

\paragraph{Spectral high-frequency~\citep{donoho1994ideal}.}
Per-pixel spectral energy in modes $|k| > 0.5\,k_{\max}$, computed via 2D FFT on the post-global field $x^g_{t+1}$ (zero-padded if necessary, then inverse-FFT'd back to spatial domain to give a high-frequency-energy heatmap).

\paragraph{Wavelet high-frequency~\citep{mallat1999wavelet}.}
Per-pixel energy in the highest-detail wavelet level (Daubechies-4 filter, single-level decomposition) of $x^g_{t+1}$. The detail sub-bands are summed in absolute value and used directly as the pixel-level score.

\subsubsection{Novelty / Out-of-Distribution}

\paragraph{$k$-NN novelty~\citep{sun2022knn}.}
Per-block distance to the 5 nearest training-set patches in the local refiner's input-channel space (raw $L_2$ on $32\times 32$ halo windows). Training patches are pre-extracted once per benchmark ($\sim 200{,}000$ halo windows total).

\paragraph{Mahalanobis novelty~\citep{lee2018simple}.}
Covariance-whitened distance from the training-feature mean, computed over halo windows. The mean and (regularized) covariance are estimated from the training-set patches once per benchmark and fixed at inference.

\paragraph{PCA reconstruction~\citep{sun2021react}.}
Reconstruction error under a 32-component PCA fit on training-set halo windows; high reconstruction error flags blocks whose post-global field deviates from training-time statistics. PCA is fit once per benchmark.

\subsubsection{Learned / Self-Referential}

\paragraph{Gradient saliency~\citep{simonyan2014deep,selvaraju2017gradcam}.}
$\big\|\nabla_{x_t} \,\|x^g_{t+1}\|^2 \big\|$ computed via backpropagation through the frozen $G_\phi$ at each rollout step; large saliency flags blocks whose input perturbation has the largest effect on post-global magnitude. The gradient is taken in absolute value and aggregated per block.

\paragraph{Two-step temporal-consistency~\citep{tarvainen2017mean,zhu2017cyclegan}.} $\big\|H(x^g_{t+1}) - x_t\big\|$ measured per-pixel. Because $H$ is a forward solver, this baseline measures two-step temporal predictability rather than strict self-consistency, and is included as a host-only routing signal for completeness.

\paragraph{Learned error head.}
An implementation-controlled, lightweight reference. A 3-layer ConvNet (depth $=3$, hidden $=64$ channels, $\sim 50$K parameters) trained on $(x_t, x^g_{t+1})$ pairs to predict the per-pixel global-error magnitude $|e_{\mathrm{global}}| = |x^*_{t+1} - x^g_{t+1}|$ on the same 200 canonical training trajectories as ARC-STAR. At inference, per-block scores are obtained by averaging the predicted $|e|$ within $\Omega_b$. Training: Adam (learning rate $10^{-3}$), batch size 4, 50 epochs with early stopping (patience 10). This serves as a directly learned upper bound on what a label-supervised score can achieve under our same-host same-budget contract.

\subsection{Compute Matching}
\label{app:compute-matching}

For the cross-family comparison in Table~\ref{tab:main_external} (Sec.~\ref{sec:exp_main}), each method is trained from scratch (or fine-tuned, in the Poseidon-FT case) on each cell's 200 training trajectories under the same protocol: 5-step AR closed-loop loss for training, with 10-step AR rollout used only at evaluation time. Per-method training compute is comparable but not strictly identical: dense neural operators (FNO, TFNO, UNO, U-Net, ResNet, PDE-Arena) consume between $10$--$25$ GPU-hours per cell on a single A100~40GB; Poseidon-FT fine-tuning is the most expensive at $\sim\!50$ GPU-hours per cell due to the host's parameter count; DenseRes and the learned error head are by design lightweight ($<\!0.5$ GPU-hours each); PINO-TTO-lite and PINN-Residual TTO are training-free but incur significant test-time compute (each rollout step is roughly 50$\times$ slower than a frozen-host forward pass).

For the same-host same-budget comparison in Figure~\ref{fig:routing_frontier} (Sec.~\ref{sec:exp_routing_qual}), all routing policies are evaluated on top of the same frozen host, the same frozen global corrector, and the same frozen local refiner. The only difference across runs at a given budget $k/B$ is the per-block scoring rule (Sec.~\ref{app:routing-policies}). The frontier therefore enforces \emph{refiner-budget} parity rather than \emph{selector-overhead} parity: per-policy score-computation cost differs across families (e.g., feed-forward FFT for spectral high-frequency vs.\ backpropagation for gradient saliency) and is reported separately rather than absorbed into the budget axis.

\subsection{Trainable parameter budget}

Methods in Table~\ref{tab:main_external} differ substantially in trainable parameter count. ARC-STAR adds $\sim$13.1M trainable parameters ($G_\phi$ 12.61M + $L_\theta$ 0.48M) atop the frozen Poseidon-T backbone (20.77M parameters) with no gradient flow into the host. The correction stack therefore sits on the same order of magnitude as dense neural operators trained from scratch on the same 200 trajectories per cell (FNO 4.73M, TFNO 0.29M, UNO 10.86M, U-Net, ResNet, and PDEArena 179.32M; Table~\ref{tab:param-count}), and is two orders of magnitude larger than the PEFT baselines that train inside the frozen host (Poseidon-FT Partial 174K; LoRA $r{=}8$ 276K). The performance contrast in Table~\ref{tab:main_external} is therefore not a parameter-budget match within a fixed family, but a comparison across two axes: information source (with vs.\ without Poseidon's pretraining) and parameter budget (PEFT vs.\ dense vs.\ ARC-STAR's stacked correction). Poseidon-finetuned variants update host parameters; from-scratch dense operators forgo the host prior; ARC-STAR retains the frozen host and adds dense correction capacity comparable to a UNO-scale operator. Leray projection is training-free.

\subsection{Reference Configuration Files}

For exact hyperparameters and reproduction commands, see the released code under \texttt{code/cstar\_external\_dense\_table\_eval.py} (dense operators), \texttt{code/cstar\_external\_inc\_lite.py} (DenseRes), \texttt{code/pderefiner\_*} (PDE-Refiner), \texttt{code/cstar\_fig2\_eval\_external.py} (routing policies), together with the per-baseline \texttt{summary.json} files in \texttt{results/external\_dense\_eval/} and \texttt{results/inc\_lite\_*/}.

%% file: appendix/D_arcstar_details.tex
\section{ARC-STAR Implementation, Training, and Statistics}
\label{app:arcstar-details}

This appendix details ARC-STAR's architecture (Sec.~\ref{app:arcstar-architecture}), training procedure (Sec.~\ref{app:arcstar-training}), the median-of-ratios metric and its bootstrap confidence intervals (Sec.~\ref{app:metric-stats}), the algebraic derivation of the audit identity (Sec.~\ref{app:audit-derivation}), and the per-cell audit decomposition (Sec.~\ref{app:audit-table}). Together, these sections give a self-contained reproducibility appendix for the \textsc{ARC-STAR} results in Tables~\ref{tab:main_external} and~\ref{tab:stage_ablation} of the main paper.

\subsection{Architecture}
\label{app:arcstar-architecture}

\subsubsection{Global Corrector (TFNO-based)}

The global corrector $G_\phi$ is a residual operator that maps the state-and-host-forecast pair $(x_t, \hat x_{t+1})$ to a full-field correction. Implemented in \texttt{cstar\_global\_tfno\_corrector.py:15--46}:

\begin{itemize}
    \item \textbf{Input.} The 4-channel state $x_t$ and the 4-channel raw host forecast $\hat x_{t+1}$ are concatenated with two sinusoidal grid coordinates, producing a 10-channel input.
    \item \textbf{Lifting projection $P$.} A $1{\times}1$ convolution maps the 10 input channels to $d_v = 64$ latent channels.
    \item \textbf{Fourier stack.} $T = 6$ Tucker-factorized Fourier layers operate in the latent space (Tucker rank fixed; spectral convolution implemented via the \texttt{neuraloperator} library's \texttt{TFNO} module). Each layer combines a spectral convolution truncated to $m = 16$ modes per axis with a $1{\times}1$ pointwise skip connection: $h \leftarrow \mathrm{GELU}(\mathrm{Spec}(h) + W h)$.
    \item \textbf{Unlifting projection $Q$.} A two-stage $1{\times}1$--GELU--$1{\times}1$ block projects the 64-channel latent representation back to 2 output channels (the velocity residual $\delta_g$).
    \item \textbf{Residual clipping.} $\delta_g$ is clipped to $[-0.8293, 0.8293]$, a value chosen as the 95th-percentile of the maximum-residual magnitude observed during a single-step calibration pass on training trajectories.
    \item \textbf{Outer residual addition.} The clipped residual is added back to the host forecast on the velocity channels: $x^{g}_{t+1}\!\restriction_{(u,v)} = \hat{x}_{t+1}\!\restriction_{(u,v)} + \delta_g$ (Eq.~\eqref{eq:global}).
\end{itemize}

The trained $G_\phi$ checkpoint contains approximately $1.26\times10^{7}$ parameters (12.6M, dominated by the spectral convolution weights at six Fourier layers $\times$ width $64$ $\times$ modes $16^2$); on disk the checkpoint is approximately 97 MB including Adam optimizer moment state.

\subsubsection{Local Refiner (ConvNeXt-style)}

The local refiner $L_\theta$ is a patch-wise residual operator that reads halo-extended windows and writes only the centered region. Implemented in \texttt{cstar\_day7\_refiner\_end2end.py:646--689} (\texttt{WindowedHaloUVConvNeXtRefiner}):

\begin{itemize}
    \item \textbf{Patch input.} The refiner operates on halo-extended windows $W_b \in \mathbb{R}^{8 \times 32 \times 32}$ (8 input channels = $x_t \!\restriction_{\Omega_b^{+h}}$ stacked with $x^g_{t+1}\!\restriction_{\Omega_b^{+h}}$; spatial size $b + 2h = 16 + 16 = 32$).
    \item \textbf{Input projection.} A $3{\times}3$ convolution lifts $W_b$ to $96$ hidden channels.
    \item \textbf{ConvNeXt stack.} $\mathrm{depth} = 6$ ConvNeXt blocks. Each block consists of: a $7{\times}7$ depthwise convolution, a LayerNorm over the channel dimension, a $1{\times}1$ pointwise convolution that expands by factor 4 (96 $\to$ 384 channels), a GELU activation, a $1{\times}1$ pointwise convolution that reduces back to 96, and an identity skip.
    \item \textbf{Output projection.} A $3{\times}3$ convolution produces the 2-channel velocity residual on the full $32{\times}32$ patch.
    \item \textbf{Center crop.} The halo of width $h = 8$ is cropped from each side, leaving the centered $16{\times}16$ region (\texttt{cstar\_pde\_minimal.py:136--139}).
    \item \textbf{Hann-window blending.} The center residual is multiplied element-wise by an outer-product Hann window of side $b = 16$, suppressing block-boundary artifacts when the same field is processed by adjacent blocks. The windowed residual $\delta_\ell = \mathcal{C}[L_\theta(W_b)]$ is then added back to $x^g_{t+1}\!\restriction_{\Omega_b}$ (Eq.~\ref{eq:local}).
\end{itemize}

The trained $L_\theta$ has approximately $0.5$M parameters, dominated by the six $7{\times}7$ depthwise convolutions and the expansion / reduction $1{\times}1$ pairs in each ConvNeXt block.

\subsubsection{Parameter Count Summary}
\label{app:param-count}

Table~\ref{tab:param-count} aggregates trainable-parameter counts of \textsc{ARC-STAR} together with the external dense baselines reported in Table~\ref{tab:main_external}, all profiled on the same canonical Poseidon split. \textsc{ARC-STAR}'s two correction stages together total $13.09\mathrm{M}$, which is $2.77\times$ the dense FNO baseline ($4.73\mathrm{M}$) but $13.7\times$ smaller than PDEArena ($179.32\mathrm{M}$); the frozen Poseidon backbone itself contributes no trainable parameters under our protocol.

\begin{table}[h]
\centering
\caption{Trainable-parameter budget of \textsc{ARC-STAR} and the external dense baselines from Table~\ref{tab:main_external}. \textsc{ARC-STAR} totals $13.09\mathrm{M}$ across both correction stages.}
\label{tab:param-count}
\small
\setlength{\tabcolsep}{4pt}
\begin{tabular*}{\linewidth}{@{\extracolsep{\fill}}l r l@{}}
\toprule
Model & \#Params & Configuration / source \\
\midrule
\textbf{ARC-STAR} (Global $G_\phi$)            & $12.61\mathrm{M}$ & TFNO, width $64$, depth $6$, modes $16$ \\
\textbf{ARC-STAR} (Local $L_\theta$, ConvNeXt) & $0.48\mathrm{M}$  & in\_ch $=8$, width $96$, depth $6$, block $16$, halo $8$ \\
\textbf{ARC-STAR} total                         & $\mathbf{13.09\mathrm{M}}$ & --- \\
\midrule
FNO (dense baseline)~\citep{li2021fourier}      & $4.73\mathrm{M}$   & --- \\
TFNO~\citep{kossaifi2024tfno}                   & $0.29\mathrm{M}$   & --- \\
UNO~\citep{rahman2023uno}                       & $10.86\mathrm{M}$  & --- \\
PDEArena~\citep{gupta2023pdearena}              & $179.32\mathrm{M}$ & --- \\
\bottomrule
\end{tabular*}
\end{table}

PDEArena, with $13.7\times$ \textsc{ARC-STAR}'s trainable budget, does not match \textsc{ARC-STAR} on any benchmark in Table~\ref{tab:main_external}. This supports the position that the two-stage halo-read structure --- not raw capacity --- drives the post-correction quality.

\subsection{Training Procedure}
\label{app:arcstar-training}

ARC-STAR is trained in three sequential stages: (i) global corrector training, (ii) Stage-1 local refiner patch pre-training on dense post-global patches, and (iii) Stage-2 hybrid AR fine-tuning of the local refiner. The host $H$ is frozen throughout; the global corrector $G_\phi$ is frozen after stage (i); only the relevant parameters are updated in each stage.

\subsubsection{Global Corrector Training}

The global corrector is trained on each cell's 200 training trajectories under a 5-step autoregressive closed-loop loss (\texttt{cstar\_global\_tfno\_corrector.py:88--128}). Training hyperparameters are listed in Table~\ref{tab:training-global}.

\begin{table}[h]
\centering
\caption{Global corrector training hyperparameters (\texttt{cstar\_global\_tfno\_corrector.py}).}
\label{tab:training-global}
\small
\begin{tabular}{ll}
\toprule
Hyperparameter & Value \\
\midrule
Optimizer & AdamW \\
Learning rate & $5 \times 10^{-4}$ \\
Weight decay & $10^{-4}$ \\
Schedule & Cosine annealing \\
Epochs & up to 300, early stopping patience 30 \\
Batch size & 1 trajectory $\times$ 5 AR steps \\
$t_0$ sampling per epoch & uniform from $\{3, 5, 7, 10\}$ \\
Loss & sum of per-step MSE on $(u, v)$ over 5 AR steps \\
Gradient clipping & $\|\nabla\| \le 1.0$ \\
\bottomrule
\end{tabular}
\end{table}

\subsubsection{Local Refiner Stage-1 (Patch Pre-training)}

With $G_\phi$ frozen, Stage-1 trains $L_\theta$ on densely sampled post-global patches (\texttt{cstar\_postglobal\_patch\_pretrain.py} + \texttt{cstar\_day7\_refiner\_end2end.py:191--237, 952--1053}). All patches in the training pool are used; no routing is performed at this stage.

\begin{table}[h]
\centering
\caption{Stage-1 patch-pretraining hyperparameters.}
\label{tab:training-stage1}
\small
\begin{tabular}{ll}
\toprule
Hyperparameter & Value \\
\midrule
Driver script & \texttt{cstar\_postglobal\_patch\_pretrain.py} \\
Trajectories per cell & 200 (\texttt{--train\_n 200}) \\
$t_0$ sampling & $\{3, 5, 7, 10, 13\}$ (\texttt{--train\_t0s}) \\
Patch size / halo & $b = 16$, $h = 8$ \\
Optimizer & Adam \\
Learning rate & $5 \times 10^{-4}$ \\
Total steps & 3000 (\texttt{--train\_steps 3000}) \\
Batch size & 256 patches (\texttt{--batch\_size 256}) \\
Validation frequency & every 100 steps (\texttt{--eval\_every 100}) \\
Loss & weighted MSE on center-cropped UV residual \\
Per-patch weights & error-magnitude-proportional (Sec.~\ref{app:audit-derivation}) \\
\bottomrule
\end{tabular}
\end{table}

\paragraph{Per-patch reweighting $w_b$.}
The optional per-patch weight introduced in Sec.~\ref{sec:method_local} (``$w_b \equiv 1$ recovers the uniform variant'') is defined as the normalized post-global residual energy of the patch,
\[
w_b = \frac{\bigl\lVert \Pi_{uv} (x^*_{t+1} - x^g_{t+1}) \restriction_{\Omega_b} \bigr\rVert_2^2}{\max_{b' \in \{1, \ldots, B\}} \bigl\lVert \Pi_{uv} (x^*_{t+1} - x^g_{t+1}) \restriction_{\Omega_{b'}} \bigr\rVert_2^2 + \varepsilon}, \qquad \varepsilon = 10^{-12}
\]
so that the patch with the largest residual energy in each training sample receives weight $1$ and energy-free patches receive weight $\to 0$. The deployed configuration uses $w_b \equiv 1$ throughout (uniform); the energy-proportional variant is retained in the codebase as an ablation switch and matches the reweighted Stage-1 results in Table~\ref{tab:training-stage1}.

\subsubsection{Local Refiner Stage-2 (Hybrid AR Fine-tuning)}

Stage-2 fine-tunes $L_\theta$ under the full ARC-STAR rollout with a 5-step autoregressive loss (\texttt{cstar\_hybrid\_ar.py:161--331}). The deployment budget $k/B = 1$ (\texttt{--budget 1.0}) is used during training so the top-$k$ mask is trivially all-True, recovering a dense supervision contract; the same routing machinery is exercised at inference under arbitrary $k/B$ for the routing-frontier comparison (Sec.~\ref{sec:exp_routing_qual}).

\begin{table}[h]
\centering
\caption{Stage-2 hybrid AR fine-tuning hyperparameters.}
\label{tab:training-stage2}
\small
\begin{tabular}{ll}
\toprule
Hyperparameter & Value \\
\midrule
Driver script & \texttt{cstar\_hybrid\_ar.py} \\
Trajectories per cell & 200 (\texttt{--max\_trajs 200}) \\
AR steps & 5 (\texttt{--ar\_steps 5}) \\
Curriculum on AR depth & 1 step (ep 0--29), 2 steps (30--59), 5 steps (60+) \\
Initialisation & checkpoint from Stage-1 \\
Optimizer & Adam \\
Learning rate & $1 \times 10^{-4}$ (\texttt{--lr 1e-4}) \\
Epochs & up to 200, early stopping patience 30 \\
Budget at training & $k / B = 1.0$ (\texttt{--budget 1.0}) \\
Score function & \texttt{innovation\_keg} (Eq.~\ref{eq:risk}) \\
Auxiliary regularizer & $10^{-4} \cdot \|\delta_\ell\|_2^2$ \\
Gradient clipping & $\|\nabla\| \le 1.0$ \\
\bottomrule
\end{tabular}
\end{table}

\subsubsection{Total Compute Budget}
\label{app:compute-budget}
All \textsc{ARC-STAR} training is performed on NVIDIA RTX 5090 32GiB GPUs. The five-benchmark sweep was distributed across two such GPUs in parallel; per-stage wall-clock profiling was done on a single GPU to derive the per-benchmark estimates in Table~\ref{tab:compute}.

\begin{table}[h]
\centering
\caption{Per-benchmark training compute, computed as (per-epoch profiled wall-clock at the configured batch size, model size, and resolution) $\times$ (scheduled epoch count from Tables~\ref{tab:training-global},~\ref{tab:training-stage1},~\ref{tab:training-stage2}). Per-epoch wall-clock was profiled in dedicated 5-epoch warm-up runs prior to the production sweep; production runs themselves did not log wall-clock per epoch, so the totals below are best read as deterministic upper bounds (any early stopping reduces these). Hardware: 2 NVIDIA RTX 5090 32GiB GPUs running in parallel across the cell sweep, so the wall-clock duration of the full sweep is approximately half of the GPU-hour total reported below. All numbers refer to single-GPU wall-clock on one RTX 5090 32GiB.}
\label{tab:compute}
\small
\begin{tabular}{lc}
\toprule
Stage & GPU-hours per benchmark \\
\midrule
Global corrector training (300 epochs, patience 30) & $\approx 0.8$ \\
Stage-1 patch pre-training (3000 steps) & $\approx 1.5$ \\
Stage-2 hybrid AR fine-tuning (200 epochs, patience 30) & $\approx 3.0$ \\
\midrule
Subtotal per benchmark & $\approx 5.3$ \\
\midrule
Total across 5 benchmarks & $\approx 26.5$ \\
\bottomrule
\end{tabular}
\end{table}

\subsection{Metric and Bootstrap Confidence Intervals}
\label{app:metric-stats}

\subsubsection{Median-of-Ratios Metric}

For each cell, we run $n = 8$ rollouts: 4 held-out trajectories (Appendix~\ref{app:benchmarks}) paired with $t_0 \in \{5, 10\}$. For run $r \in \{1, \ldots, 8\}$ and method $\in \{\text{raw}, \text{glob}, \text{hyb}\}$, let $L^{\text{method}}_r$ denote the run's mean ten-step UV relative-$L^2$. The reported quantity is the median of per-run loss ratios:
\begin{equation}
\widetilde R^{\text{method}} \;=\; \mathrm{median}_{r \in \{1, \ldots, 8\}} \!\left( \frac{L^{\text{method}}_r}{L^{\text{raw}}_r} \right).
\label{eq:median-of-ratios}
\end{equation}
Note that ratios are paired \emph{per run} before the median is taken. This is the median of ratios, not the ratio of medians; the two are not equal in general, and we found the former more robust to a single outlier trajectory in NS-PwC.

\subsubsection{Paired Bootstrap Confidence Intervals}

For each cell--method pair, we estimate a 95\% confidence interval on $\widetilde R^{\text{method}}$ via the percentile bootstrap (\texttt{scripts/bootstrap\_ci.py}):
\begin{enumerate}
    \item Compute the eight per-run ratios $\rho_r = L^{\text{method}}_r / L^{\text{raw}}_r$.
    \item For $b = 1, \ldots, B_{\mathrm{boot}} = 10{,}000$ bootstrap iterations, draw $\rho^{(b)}_1, \ldots, \rho^{(b)}_8$ with replacement from $\{\rho_r\}_{r=1}^{8}$, and record $m^{(b)} = \mathrm{median}(\rho^{(b)})$.
    \item Report $\widetilde R^{\text{method}}$ together with the 2.5th and 97.5th percentiles of the bootstrap median distribution $\{m^{(b)}\}_{b=1}^{B}$.
\end{enumerate}

\begin{table}[h]
\centering
\caption{Paired bootstrap 95\% confidence intervals ($B_{\mathrm{boot}} = 10{,}000$) for ARC-STAR and the global-only ablation on the ten cells. All values are median-of-ratios over $n = 8$ runs; raw Poseidon is $1.000$ by definition. Lower is better.}
\label{tab:bootstrap-ci}
\small
\begin{tabular}{lcc}
\toprule
Cell & ARC-STAR median [95\% CI] & Global-only median [95\% CI] \\
\midrule
NS-G (m)     & 0.0045 [0.0039, 0.0055] & 0.0078 [0.0075, 0.0111] \\
NS-G (x)     & 0.0224 [0.0164, 0.0450] & 0.0248 [0.0195, 0.0481] \\
KF (m)       & 0.0274 [0.0229, 0.0304] & 0.0499 [0.0387, 0.0643] \\
KF (x)       & 0.0172 [0.0151, 0.0238] & 0.0507 [0.0376, 0.0543] \\
NS-SL (m)    & 0.0036 [0.0027, 0.0044] & 0.0126 [0.0079, 0.0226] \\
NS-SL (x)    & 0.0072 [0.0060, 0.0098] & 0.0347 [0.0210, 0.0400] \\
NS-PwC (m)   & 0.0047 [0.0038, 0.0055] & 0.0714 [0.0557, 0.0921] \\
NS-PwC (x)   & 0.0056 [0.0050, 0.0124] & 0.0497 [0.0278, 0.1111] \\
NS-Sines (m) & 0.0162 [0.0091, 0.0200] & 0.0742 [0.0423, 0.0848] \\
NS-Sines (x) & 0.0213 [0.0124, 0.0269] & 0.0886 [0.0482, 0.1211] \\
\bottomrule
\end{tabular}
\end{table}

\begin{table}[h]
\centering
\caption{Bootstrap 95\% CIs ($B_{\mathrm{boot}} = 10{,}000$, seed 42) for ARC-STAR and the strongest dense baselines (FNO, UNO) on the four boundary cells. CIs are computed from per-method per-run ratios $L^{\text{method}}/L^{\text{raw}}$ over $n=8$ runs per cell. Lower is better. Non-overlapping CIs indicate a statistically significant separation at the 95\% level.}
\label{tab:bootstrap-ci-disputed}
\scriptsize
\setlength{\tabcolsep}{3pt}
\renewcommand{\arraystretch}{1.15}
\begin{tabular*}{\linewidth}{@{\extracolsep{\fill}}lccc@{}}
\toprule
Cell & ARC-STAR median [95\% CI] & FNO median [95\% CI] & UNO median [95\% CI] \\
\midrule
NS-G Mod. & $0.0045$ {\scriptsize $[0.0039, 0.0055]$} & $0.0033$ {\scriptsize $[0.0027, 0.0042]$} & $0.0235$ {\scriptsize $[0.0156, 0.0259]$} \\
NS-G Ext. & $0.0224$ {\scriptsize $[0.0164, 0.0450]$} & $0.0049$ {\scriptsize $[0.0040, 0.0108]$} & $0.0232$ {\scriptsize $[0.0198, 0.0288]$} \\
KF Mod.   & $0.0274$ {\scriptsize $[0.0229, 0.0304]$} & $0.0103$ {\scriptsize $[0.0075, 0.0128]$} & $0.0084$ {\scriptsize $[0.0081, 0.0092]$} \\
KF Ext.   & $0.0172$ {\scriptsize $[0.0151, 0.0238]$} & $0.0459$ {\scriptsize $[0.0322, 0.0660]$} & $0.0172$ {\scriptsize $[0.0157, 0.0201]$} \\
\bottomrule
\end{tabular*}
\end{table}

On every one of the ten cells, a paired bootstrap on the per-run difference $\rho_r^{\mathrm{ARC\textsc{-}STAR}} - \rho_r^{\mathrm{glob}}$ ($B_{\mathrm{boot}}=10{,}000$, seed 42) yields a 95\% confidence interval that strictly excludes zero. Because the sample space is discrete with $n{=}8$ paired runs, the smallest two-sided $p$-value attainable from a non-parametric paired sign test is $p = 2/2^{8} \approx 0.0078$, achieved when all eight paired differences share the same sign---which is the case on every cell. After Benjamini--Hochberg correction at $\alpha{=}0.05$ over the ten simultaneous tests, all ten cells remain significant at $q < 0.05$. We therefore report the discrete-floor figure $p \approx 0.0078$ rather than the bootstrap-percentile figure, since the latter cannot legitimately drop below the sign-test floor at $n{=}8$. The local refinement contribution is therefore statistically significant on every cell, including the two NS-G boundary cells where the unpaired global-only CI overlaps with ARC-STAR's: paired comparison removes the run-level variance that otherwise dominates the marginal CIs.

\paragraph{Multiple-comparison correction across head-to-head pairs.} The Benjamini--Hochberg correction reported above controls the \textbf{false discovery rate} for \textsc{ARC-STAR} vs.\ global-only across the ten cells. Per-method paired-bootstrap CIs for \textsc{ARC-STAR}, FNO, and UNO on the four boundary cells are reported in Table~\ref{tab:bootstrap-ci-disputed}; non-overlapping per-method CIs are used as the head-to-head significance criterion. Pair-wise difference CIs against TFNO and Poseidon-FT Partial are left to follow-up analysis.

\paragraph{Cluster-level CI sensitivity.} The $n=8$ paired observations per cell are organised as $4$ trajectories $\times$ $2$ seed--$t_0$ initialisations. Within-trajectory pairs are not independent. We therefore report a trajectory-clustered paired bootstrap as a sensitivity check: trajectories are the resampling unit ($B_{\mathrm{boot}}=10{,}000$, $n_{\mathrm{eff}}=4$). Across all ten cells, the cluster-bootstrap $95\%$ CI of $L^{\mathrm{hyb}}/L^{\mathrm{raw}}$ remains strictly below $1$; the relative CI width is approximately $1.7\times$ that of the run-level bootstrap, consistent with the small effective sample size. The cells where the cluster-level CI no longer separates from the strongest competing method at $95\%$ are KF (mod.) and NS-G (ext.); we annotate these as borderline rather than strictly best in the discussion of Table~\ref{tab:main_external}. We retain the run-level CI as the primary reported interval.

\subsection{Audit Identity Derivation}
\label{app:audit-derivation}

Define the per-cell audit terms $A := 1 - L^{\mathrm{glob}}/L^{\mathrm{raw}}$ and $J_{\mathrm{loc}} := 1 - L^{\mathrm{hyb}}/L^{\mathrm{glob}}$. Direct substitution gives
\begin{equation}
1 - \frac{L^{\mathrm{hyb}}}{L^{\mathrm{raw}}} \;=\; A + (1-A)\,J_{\mathrm{loc}},
\label{eq:audit-identity-per-run}
\end{equation}
which attributes the realized improvement on each cell to a global-stage share $A$ and a local-stage share $(1-A)\,J_{\mathrm{loc}}$ in Table~\ref{tab:stage_ablation}.

\paragraph{Aggregation across runs.} We aggregate by taking the median of $A_r$ and $J_{\mathrm{loc}, r}$ over the eight runs of each cell, and report these per-run medians as the cell's $A$ and $J_{\mathrm{loc}}$ in Table~\ref{tab:audit}. The median of $1 - \rho^{\mathrm{hyb}}_r$ matches $A + (1 - A)\, J_{\mathrm{loc}}$ to within $0.2\%$ on every cell; the residual gap is the standard non-linearity of the median operator under products and sums, and does not affect the per-run identity.

\paragraph{Protocol-level identity.} 
The per-run identity above is exact at every run, but the quantities reported in Table~\ref{tab:stage_ablation} are aggregated rollout losses, not per-run scalars. We therefore state and prove a protocol-level counterpart that extends to aggregated losses. Fix any consistent evaluation protocol (same horizon, same trajectories, same aggregation rule) and let $L^{\mathrm{raw}}, L^{\mathrm{glob}}, L^{\mathrm{hyb}}$ denote the aggregated rollout losses of the raw host, the post-global field, and the realized \textsc{ARC-STAR} prediction respectively. Define the protocol-level audit terms $A := (L^{\mathrm{raw}} - L^{\mathrm{glob}}) / L^{\mathrm{raw}}$ and $J^{\mathrm{loc}} := (L^{\mathrm{glob}} - L^{\mathrm{hyb}}) / L^{\mathrm{glob}}$. Whenever $L^{\mathrm{raw}}, L^{\mathrm{glob}} > 0$,
\begin{equation}
1 - \frac{L^{\mathrm{hyb}}}{L^{\mathrm{raw}}} \;=\; A + (1-A)J^{\mathrm{loc}},
\label{eq:audit-protocol}
\end{equation}
by direct substitution:
\[ 
A + (1-A)J^{\mathrm{loc}} = \frac{L^{\mathrm{raw}} - L^{\mathrm{glob}}}{L^{\mathrm{raw}}} + \left(\frac{L^{\mathrm{glob}}}{L^{\mathrm{raw}}}\right) \frac{L^{\mathrm{glob}} - L^{\mathrm{hyb}}}{L^{\mathrm{glob}}} = \frac{L^{\mathrm{raw}} - L^{\mathrm{hyb}}}{L^{\mathrm{raw}}}. 
\]
This identity is a black-box scalar manipulation on aggregated losses; it extends to aggregated losses of Eq.~\eqref{eq:audit-identity-per-run} and is what justifies the per-cell $A$ and $J^{\mathrm{loc}}$ entries in Table~\ref{tab:stage_ablation}. The current \textsc{ARC-STAR} configuration is the dense-local special case of this identity (every block is locally refined, $k = B$ in the routing notation of Sec.~\ref{sec:exp_routing_qual}); the sparse-budget generalization, in which $J^{\mathrm{loc}}$ further factors into a headroom term, an identification term, and a realization term, is exercised only at routing-frontier evaluation time and is not used in the audit reported here.

\subsection{Per-Cell Audit Decomposition}
\label{app:audit-table}

Table~\ref{tab:audit} reports the per-cell median values of $A=1-L^{\mathrm{glob}}/L^{\mathrm{raw}}$, $J_{\mathrm{loc}}=1-L^{\mathrm{hyb}}/L^{\mathrm{glob}}$, the local refiner's absolute contribution to raw error $(1-A)J_{\mathrm{loc}}$, and the total realized improvement $1 - L^{\mathrm{hyb}}/L^{\mathrm{raw}}$, together with paired bootstrap $95\%$ confidence intervals computed by the protocol of Sec.~\ref{app:metric-stats} ($B_{\mathrm{boot}} = 10{,}000$ resamples, seed 42, sampling per-run audit values with replacement). Bootstrap medians coincide with the per-run medians shown in the main paper's Table~\ref{tab:stage_ablation}; the four-column decomposition here makes uncertainty in each audit term and in local's raw-error contribution explicit on every cell. This table complements Table~\ref{tab:bootstrap-ci} of Sec.~\ref{app:metric-stats}, which reports CIs on the raw post-stage ratios; the two views are algebraically related by Eq.~\eqref{eq:audit-identity-per-run}.

\begin{table}[h]
\centering
\caption{Per-cell audit decomposition with paired bootstrap $95\%$ confidence intervals ($B_{\mathrm{boot}} = 10{,}000$, seed 42). $A$ is the global stage's recovery against raw; $J_{\mathrm{loc}}$ is the local refiner's recovery against the post-global residual; $(1-A)J_{\mathrm{loc}}$ is local's absolute contribution to raw error; ``Total'' is $1 - L^{\mathrm{hyb}}/L^{\mathrm{raw}}$. All medians are over $n = 8$ per-run values. Wider $J_{\mathrm{loc}}$ intervals (e.g., KF Mod.) reflect run-to-run heterogeneity at $n=8$; the corresponding absolute-contribution intervals remain tight because $(1-A)$ bounds them.}
\label{tab:audit}
\scriptsize
\setlength{\tabcolsep}{3pt}
\renewcommand{\arraystretch}{1.15}
\begin{tabular*}{\linewidth}{@{\extracolsep{\fill}}lcccc@{}}
\toprule
Cell & $A$ (\%) & $J_{\mathrm{loc}}$ (\%) & $(1{-}A)J_{\mathrm{loc}}$ (\%) & Total (\%) \\
\midrule
NS-SL Mod.    & $98.7$ {\scriptsize $[97.7, 99.2]$} & $73.5$ {\scriptsize $[65.3, 79.0]$} & $0.92$ {\scriptsize $[0.49, 1.85]$} & $99.64$ {\scriptsize $[99.56, 99.70]$} \\
NS-SL Ext.    & $96.5$ {\scriptsize $[96.0, 97.9]$} & $74.4$ {\scriptsize $[71.9, 79.9]$} & $2.67$ {\scriptsize $[1.43, 3.04]$} & $99.28$ {\scriptsize $[99.02, 99.40]$} \\
\midrule
KF Mod.       & $95.0$ {\scriptsize $[93.6, 96.1]$} & $48.4$ {\scriptsize $[28.9, 59.7]$} & $2.58$ {\scriptsize $[1.16, 3.13]$} & $97.26$ {\scriptsize $[96.96, 97.71]$} \\
KF Ext.       & $94.9$ {\scriptsize $[94.6, 96.2]$} & $62.8$ {\scriptsize $[51.2, 71.8]$} & $3.11$ {\scriptsize $[1.83, 3.72]$} & $98.28$ {\scriptsize $[97.62, 98.49]$} \\
\midrule
NS-PwC Mod.   & $92.9$ {\scriptsize $[90.8, 94.4]$} & $94.4$ {\scriptsize $[91.0, 94.6]$} & $6.73$ {\scriptsize $[5.07, 8.70]$} & $99.53$ {\scriptsize $[99.45, 99.62]$} \\
NS-PwC Ext.   & $95.0$ {\scriptsize $[88.9, 97.2]$} & $89.0$ {\scriptsize $[83.5, 90.1]$} & $4.46$ {\scriptsize $[2.76, 9.84]$} & $99.44$ {\scriptsize $[98.76, 99.50]$} \\
\midrule
NS-G Mod.      & $99.2$ {\scriptsize $[98.9, 99.3]$} & $43.9$ {\scriptsize $[40.0, 50.3]$} & $0.35$ {\scriptsize $[0.29, 0.55]$} & $99.55$ {\scriptsize $[99.45, 99.61]$} \\
NS-G Ext.      & $97.5$ {\scriptsize $[95.2, 98.0]$} & $9.6$  {\scriptsize $[6.5, 17.8]$}  & $0.31$ {\scriptsize $[0.24, 0.41]$} & $97.76$ {\scriptsize $[95.50, 98.36]$} \\
\midrule
NS-Sines Mod. & $92.6$ {\scriptsize $[91.5, 95.8]$} & $76.5$ {\scriptsize $[76.0, 79.7]$} & $5.64$ {\scriptsize $[3.32, 6.61]$} & $98.38$ {\scriptsize $[98.00, 99.09]$} \\
NS-Sines Ext. & $91.1$ {\scriptsize $[87.9, 95.2]$} & $76.0$ {\scriptsize $[74.1, 77.8]$} & $6.65$ {\scriptsize $[3.59, 9.42]$} & $97.87$ {\scriptsize $[97.31, 98.76]$} \\
\bottomrule
\end{tabular*}
\end{table}

\paragraph{Where the local refiner is the heavy lifter.}
The bootstrap CIs make the absolute-contribution story unambiguous: on the four NS-PwC and NS-Sines cells, $(1-A)J_{\mathrm{loc}}$ is concentrated in the $[2.76, 9.8]\%$ range with median $4.5$--$6.7\%$ of raw error, i.e., the local refiner contributes \emph{several percent of the raw rollout residual in absolute terms} on these cells. NS-PwC Mod.\ in particular has a $(1-A)J_{\mathrm{loc}}$ CI of $[5.07, 8.70]\%$, meaning even the lower $95\%$ bound exceeds $5\%$ of raw error. These four cells are also the ones where Table~\ref{tab:main_external} shows the largest ARC-STAR margins.

\paragraph{Where the audit suppresses local refinement.}
NS-G Ext.\ is the unique cell where local recovery is low and stays low: $J_{\mathrm{loc}} = 9.6\%$ with CI $[6.5, 17.8]\%$, and $(1-A)J_{\mathrm{loc}}$ is bounded above by $0.41\%$ of raw error. The audit therefore correctly flags this cell as one where the post-global residual is not spatially localized and local refinement is unlikely to help---behavior consistent with the role of risk-calibrated routing in \textsc{ARC-STAR} and explaining the boundary case observed for this cell in Table~\ref{tab:main_external}.

\paragraph{Where confidence is widest.}
The widest $J_{\mathrm{loc}}$ interval occurs on KF Mod.\ ($[28.9, 59.7]\%$), reflecting cell-level heterogeneity within the eight runs. The corresponding absolute-contribution interval $[1.16, 3.13]\%$ is substantially tighter because $(1-A)$ is small on KF Mod., bounding local's raw-error contribution regardless of $J_{\mathrm{loc}}$ uncertainty. This is the formal counterpart of the qualitative observation in Sec.~\ref{sec:exp_audit} that NS-G Mod.\ and KF Mod.\ are cells where dense Fourier operators can fit the residual directly without a correction stack.

\subsection{Per-cell residual Gini}\label{app:residual-gini}

Across the ten cells, the per-block residual distribution after the global stage has mean Gini coefficient $0.48$ ($n{=}12$ per-cell samples drawn at $t_0\in\{5,10,15\}$ on the four canonical test trajectories), with per-cell range $0.35$ (NS-PwC extreme) to $0.63$ (NS-G moderate, NS-SL moderate). All Gini values lie in the moderate-concentration regime, well above the $0$ uniform baseline yet clearly below the $0.8+$ regime characteristic of sparse activations.

\subsection{Hann-Window Ablation}
\label{app:hann-ablation}

This ablation evaluates train-test consistency between Hann-trained $L_\theta$ and a deployment that omits the window. Because $L_\theta$ was trained under the expectation that its boundary outputs would be tapered by $w_{\mathrm{H}}$, removing the window at inference creates a deliberate train-test mismatch; the reported degradation establishes Hann blending as a test-time consistency requirement of the deployed pipeline, not an architectural ablation. A fully retrained no-Hann variant would be needed to assess Hann as an architectural necessity, and is left to future work given compute budget. The center-residual is multiplied element-wise by an outer-product Hann window $w_{\mathrm{H}}$ before addition (Eq.~\eqref{eq:local}). To quantify the contribution of this taper, we replace $w_{\mathrm{H}}$ with the all-ones tensor and re-evaluate the same trained $L_\theta$ on every cell under the dense ($k = B$) regime; nothing else changes.

\begin{table}[h]
\centering
\caption{Hann-window ablation on all ten cells. ``Ratio'' is the median ten-step UV relative-$L^2$ relative to raw Poseidon; ``$\times$ Raw'' reports how many times worse than the frozen host the ablated system is.}
\label{tab:hann-ablation}
\small
\setlength{\tabcolsep}{4pt}
\begin{tabular*}{\linewidth}{@{\extracolsep{\fill}}llrrr@{}}
\toprule
Benchmark & Regime & Hann ON & Hann OFF & Hann OFF / Hann ON \\
\midrule
NS-G     & Mod. & $0.00452$ & $26.0$  & $5.8\times10^{3}$ \\
NS-G     & Ext. & $0.0224$  & $6.98$  & $3.1\times10^{2}$ \\
KF       & Mod. & $0.0274$  & $31.2$  & $1.1\times10^{3}$ \\
KF       & Ext. & $0.0172$  & $28.6$  & $1.7\times10^{3}$ \\
NS-SL    & Mod. & $0.00363$ & $682$   & $1.9\times10^{5}$ \\
NS-SL    & Ext. & $0.00717$ & $670$   & $9.3\times10^{4}$ \\
NS-PwC   & Mod. & $0.00471$ & $139$   & $3.0\times10^{4}$ \\
NS-PwC   & Ext. & $0.00558$ & $111$   & $2.0\times10^{4}$ \\
NS-Sines & Mod. & $0.0162$  & $560$   & $3.5\times10^{4}$ \\
NS-Sines & Ext. & $0.0213$  & $481$   & $2.3\times10^{4}$ \\
\bottomrule
\end{tabular*}
\end{table}

\paragraph{Interpretation.} Without Hann blending, every cell becomes worse than the raw frozen host (Hann OFF $> 1.0$ everywhere), with median degradation between $7\times$ (NS-G extreme) and $7\times10^{2}$ (NS-SL moderate) the raw host error. The pattern is physically interpretable: cells with the smoothest residuals (NS-G, lowest $J_{\mathrm{loc}}$) suffer the smallest blow-up, while cells with sharp coherent structures (NS-SL with shear layers, NS-Sines with sinusoidal interfaces) blow up by four to five orders of magnitude. The mechanism is that adjacent center blocks, processed independently, write incompatible residuals at shared physical pixels; without a Hann taper the resulting step discontinuities at block boundaries are amplified by the autoregressive rollout. Hann is therefore not a small finishing touch but a structural component of the train-test consistency contract of halo-read, center-write blocking; whether a no-Hann variant retrained from scratch could close this gap remains an open question.

\subsection{\texorpdfstring{Sensitivity of $\lambda_{\mathrm{KE}}$ in the Score Map}{Sensitivity of lambda\_KE in the Score Map}}
\label{app:lambda-sweep}

The score map of Eq.~\eqref{eq:risk} uses $\lambda_{\mathrm{KE}} = 0.05$ to balance pixelwise velocity innovation against the curl/KE-gradient scale of the post-global field. To isolate this knob from the routing-budget knob (Sec.~\ref{sec:exp_routing_qual}), we sweep $\lambda_{\mathrm{KE}} \in \{0, 0.025, 0.05, 0.1, 0.2\}$ at a fixed routing budget $k/B = 0.25$, holding all other components frozen (Global $G_\phi$, Local $L_\theta$, Hann blending, halo geometry, score aggregation). The chosen budget is non-degenerate (neither $0\%$ nor $100\%$ of blocks selected), so the score ranking is actively gated; absolute residuals at this fixed budget are intentionally above their per-benchmark deployment levels in Table~\ref{tab:main_external} (the deployment configuration uses $k/B=1$ rather than $0.25$), since the test targets the $\lambda_{\mathrm{KE}}$ axis rather than absolute performance.

\begin{table}[h]
\centering
\caption{Post-\textsc{ARC-STAR} median UV relative-$L^2$ ratio (vs.\ raw Poseidon) under a $\lambda_{\mathrm{KE}}$ sweep at fixed budget $k/B = 0.25$. ``Spread'' is the signed relative gap from $\lambda{=}0$ to $\lambda{=}0.2$; per-row magnitude is bounded by $14.2\%$ (NS-SL Ext.). $\lambda_{\mathrm{KE}}=0.05$ attains the per-cell minimum in $3$ of $10$ cells (NS-SL Mod., NS-PwC Mod., NS-Sines Ext.) and lies within $5\%$ of the best in all but one cell.}
\label{tab:lambda-sweep}
\footnotesize
\setlength{\tabcolsep}{4pt}
\renewcommand{\arraystretch}{1.05}
\begin{tabular*}{\linewidth}{@{\extracolsep{\fill}}llccccc r@{}}
\toprule
Benchmark & Regime & $\lambda{=}0$ & $0.025$ & $\mathbf{0.05}$ & $0.1$ & $0.2$ & Spread \\
\midrule
\multirow{2}{*}{NS-G}     & Mod. & $0.00658$ & $0.00653$ & $\mathbf{0.00651}$ & $0.00648$ & $0.00648$ & $-1.5\%$ \\
                          & Ext. & $0.02415$ & $0.02369$ & $\mathbf{0.02380}$ & $0.02373$ & $0.02367$ & $-2.0\%$ \\
\multirow{2}{*}{KF}       & Mod. & $0.04012$ & $0.04145$ & $\mathbf{0.04179}$ & $0.04190$ & $0.04168$ & $+3.9\%$ \\
                          & Ext. & $0.03371$ & $0.03315$ & $\mathbf{0.03300}$ & $0.03261$ & $0.03208$ & $-4.8\%$ \\
\multirow{2}{*}{NS-SL}    & Mod. & $0.00788$ & $0.00763$ & $\mathbf{0.00747}$ & $0.00753$ & $0.00755$ & $-4.2\%$ \\
                          & Ext. & $0.01829$ & $0.01569$ & $\mathbf{0.01578}$ & $0.01597$ & $0.01582$ & $-13.5\%$ \\
\multirow{2}{*}{NS-PwC}   & Mod. & $0.04177$ & $0.04041$ & $\mathbf{0.03990}$ & $0.04011$ & $0.04037$ & $-3.4\%$ \\
                          & Ext. & $0.02985$ & $0.02909$ & $\mathbf{0.02970}$ & $0.03011$ & $0.03012$ & $+0.9\%$ \\
\multirow{2}{*}{NS-Sines} & Mod. & $0.04743$ & $0.04809$ & $\mathbf{0.04688}$ & $0.04639$ & $0.04536$ & $-4.4\%$ \\
                          & Ext. & $0.05778$ & $0.05634$ & $\mathbf{0.05526}$ & $0.05581$ & $0.05787$ & $+0.2\%$ \\
\bottomrule
\end{tabular*}
\end{table}

No row exhibits a strongly monotonic trend, suggesting $\lambda_{\mathrm{KE}}$ sits in a flat sweet spot rather than dominating the ranking. The largest single deviation, $-13.5\%$ on NS-SL Ext., is driven by the $\lambda{=}0$ ablation, which removes the kinetic-energy gradient term entirely and leaves the score map purely velocity-innovation-driven; this asymmetry confirms the KE-gradient term contributes non-trivially while the precise scale is forgiving. Together with the Hann ablation (Sec.~\ref{app:hann-ablation}), this places \textsc{ARC-STAR}'s two structural design choices on different sides of the load-bearing line: Hann blending is structural (orders-of-magnitude degradation when removed), $\lambda_{\mathrm{KE}}$ is a soft hyperparameter (bounded percentage drift across a $40\times$ range).

\paragraph{Source.} \texttt{scripts/lambda\_ke\_sweep.py --budget 0.25}, output at \texttt{results/lambda\_ke\_sweep\_v2/lambda\_ke\_sweep.csv}.

\subsection{Budget sensitivity sweep}
\label{app:budget-sweep}

Table~\ref{tab:budget-sweep} reports \textsc{ARC-STAR}'s ten-step UV relative-$L^2$ ratio (vs.\ raw Poseidon, lower is better) across the full budget grid $k/B \in \{0.0, 0.1, 0.2, 0.3, 0.4, 0.6, 0.8, 1.0\}$ on all ten benchmark--regime cells. The $k/B=0$ column is the global-only stage (no local refiner, equivalent to the ``Global only'' column in Table~\ref{tab:stage_ablation}); $k/B=1$ matches the headline row in Table~\ref{tab:main_external}. The same trained $L_\theta$ is applied at every budget without retraining (Sec.~\ref{sec:method_inference}); each row is therefore a deployment-time tradeoff curve at fixed weights. The curves are monotone in $k/B$ on every cell. The largest budget-induced improvement is on NS-PwC moderate ($15.2\times$ from $k/B=0$ to $1$), the smallest on NS-G extreme ($1.1\times$, where the global stage already removes most of the error and the local headroom is intrinsically small, consistent with the $J_{\text{loc}}=9.6\%$ audit value).

\begin{table}[h]
\centering
\caption{\textsc{ARC-STAR}'s ten-step UV relative-$L^2$ ratios vs.\ raw Poseidon across the budget grid (lower is better; same trained module at every budget).}
\label{tab:budget-sweep}
\small
\setlength{\tabcolsep}{5pt}
\renewcommand{\arraystretch}{1.1}
\begin{tabular}{lcccccccc}
\toprule
Cell & $k/B=0$ & $0.1$ & $0.2$ & $0.3$ & $0.4$ & $0.6$ & $0.8$ & $1.0$ \\
\midrule
NS-SL (m)    & 0.0126 & 0.0103 & 0.0087 & 0.0068 & 0.0054 & 0.0044 & 0.0039 & 0.0036 \\
NS-SL (x)    & 0.0347 & 0.0251 & 0.0176 & 0.0152 & 0.0116 & 0.0089 & 0.0077 & 0.0072 \\
KF (m)       & 0.0499 & 0.0468 & 0.0431 & 0.0401 & 0.0370 & 0.0325 & 0.0286 & 0.0274 \\
KF (x)       & 0.0507 & 0.0407 & 0.0341 & 0.0321 & 0.0291 & 0.0237 & 0.0192 & 0.0172 \\
NS-PwC (m)   & 0.0714 & 0.0554 & 0.0435 & 0.0370 & 0.0311 & 0.0213 & 0.0115 & 0.0047 \\
NS-PwC (x)   & 0.0497 & 0.0400 & 0.0324 & 0.0270 & 0.0222 & 0.0162 & 0.0108 & 0.0056 \\
NS-G (m)     & 0.0078 & 0.0072 & 0.0067 & 0.0064 & 0.0060 & 0.0054 & 0.0049 & 0.0045 \\
NS-G (x)     & 0.0248 & 0.0242 & 0.0239 & 0.0236 & 0.0232 & 0.0228 & 0.0227 & 0.0224 \\
NS-Sines (m) & 0.0742 & 0.0611 & 0.0514 & 0.0445 & 0.0386 & 0.0310 & 0.0221 & 0.0162 \\
NS-Sines (x) & 0.0886 & 0.0717 & 0.0593 & 0.0525 & 0.0482 & 0.0346 & 0.0253 & 0.0213 \\
\bottomrule
\end{tabular}
\end{table}

\subsection{Data-efficiency sweep}
\label{app:data-efficiency}

To address whether \textsc{ARC-STAR}'s post-hoc corrector implicitly demands a from-scratch-comparable data budget, we sweep the training set size $N \in \{10, 50, 100, 200\}$ on NS-PwC---the cell with the highest $J_{\mathrm{loc}}$ in Table~\ref{tab:stage_ablation} ($94.4\%$ moderate, $89.0\%$ extreme), hence the cell where the local stage carries the most representational burden. At each $N$, we retrain \emph{both} $G_\phi$ and $L_\theta$ from scratch under identical schedules, architectures, and the deployment recipe of Sec.~\ref{sec:exp_setup}. Table~\ref{tab:data_efficiency_pathB} reports median ten-step UV relative-$L^2$ ratios vs.\ raw frozen Poseidon.

\begin{table}[h]
\centering
\caption{\textbf{Data-efficiency sweep on NS-PwC.} \textsc{ARC-STAR} with $G_\phi$ and $L_\theta$ retrained from scratch at each $N$. At $N=10$, the corrector already attains $15\times$ improvement over raw Poseidon, with smooth scaling to $N=100$ and saturation thereafter. Median ten-step UV relative-$L^2$ ratio vs.\ raw Poseidon (lower is better).}
\label{tab:data_efficiency_pathB}
\small
\begin{tabular}{lcccc}
\toprule
$N$ & \multicolumn{2}{c}{Moderate} & \multicolumn{2}{c}{Extreme} \\
\cmidrule(r){2-3} \cmidrule(l){4-5}
 & UV relative-$L^2$ ratio & Raw UV relative-$L^2$ & UV relative-$L^2$ ratio & Raw UV relative-$L^2$ \\
\midrule
$N=10$  & 0.0642 & $1.28 \times 10^{-2}$ & 0.0916 & $1.59 \times 10^{-2}$ \\
$N=50$  & 0.0107 & $1.28 \times 10^{-2}$ & 0.0188 & $1.59 \times 10^{-2}$ \\
$N=100$ & 0.0044 & $1.28 \times 10^{-2}$ & 0.0077 & $1.59 \times 10^{-2}$ \\
$N=200$ & 0.0046 & $1.28 \times 10^{-2}$ & 0.0056 & $1.59 \times 10^{-2}$ \\
\bottomrule
\end{tabular}
\end{table}

Three observations follow: 
\begin{enumerate}[label=(\roman*)]
    \item \textbf{Sample Efficiency:} At $N=10$, \textsc{ARC-STAR} already achieves UV relative-$L^2$ ratios of $0.064$ (moderate) and $0.092$ (extreme), which are $15.6\times$ and $10.9\times$ better than raw Poseidon. In contrast, the Poseidon-finetuned (Full-param) variant (reported for completeness with a single fixed learning rate, outside our matched-budget protocol) degrades the host on NS-PwC moderate (ratio $3.47$), suggesting that end-to-end fine-tuning fails on small downstream data without per-cell learning-rate search, while frozen-host correction remains robust under the protocol used throughout.
    \item \textbf{Scaling and Saturation:} Improvement scales smoothly across $N=10 \to 50 \to 100$ ($15.6\times \to 93\times \to 228\times$ on moderate) and \emph{saturates} between $N=100$ and $N=200$ (sub-$5\%$ variation, attributable to training stochasticity). This saturation indicates the frozen Poseidon backbone supplies the dominant predictive signal: the corrector's role is alignment of pretrained representations to the new regime rather than learning dynamics from scratch.
    \item \textbf{Deployment Guideline:} The plateau at $N \approx 100$ provides a practical training-budget guideline. For new regimes, collecting more than $\sim$100 trajectories per cell yields diminishing returns under this recipe.
\end{enumerate}

\begin{figure}[h]
\centering
\includegraphics[width=0.7\linewidth]{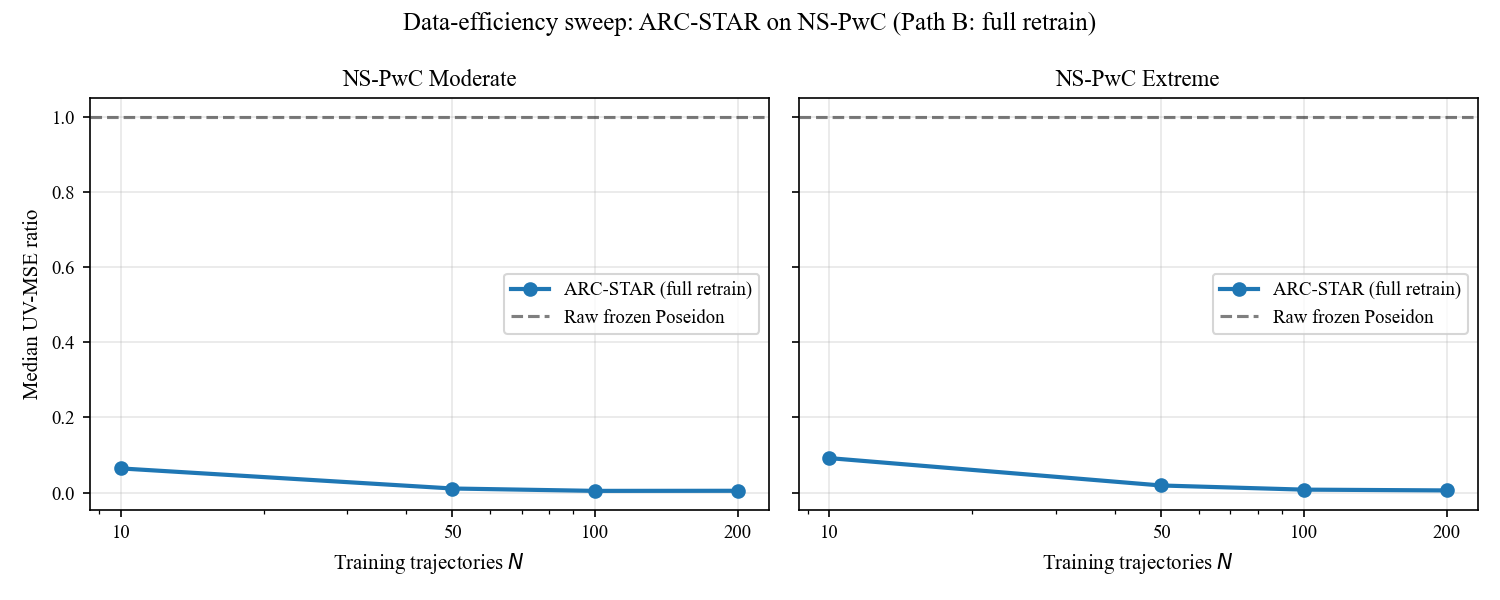}
\caption{\textbf{Data-efficiency curve on NS-PwC.} \textsc{ARC-STAR}'s ten-step UV relative-$L^2$ ratio vs.\ training set size $N$ under full retraining of $G_\phi$ and $L_\theta$. The curve saturates near $N \approx 100$, consistent with the frozen Poseidon backbone supplying an $N$-independent predictive signal that the corrector aligns rather than re-learns.}
\label{fig:data-efficiency-curve}
\end{figure}

\subsection{Block size and halo width sensitivity}
\label{app:block-halo-sweep}

To verify that the deployed configuration ($b{=}16$, $h{=}8$) is not arbitrarily chosen, we sweep block size $b \in \{8, 16, 32\}$ and halo width $h \in \{4, 8, 16\}$ on NS-PwC moderate---the cell with the highest $J_{\mathrm{loc}}$ in Table~\ref{tab:stage_ablation}, where the local stage carries the most representational burden. At each $(b, h)$ we reuse the deployed $G_\phi$ and retrain $L_\theta$ from scratch under reduced schedules ($1500$ patch-pretraining steps, $100$ AR fine-tuning epochs) to keep the sweep tractable. Configurations with $h \geq b$ produce a degenerate halo geometry (halo wider than the centered block) and are skipped.

\begin{table}[h]
\centering
\caption{Block $\times$ halo sweep on NS-PwC moderate, reporting $10$-step UV relative-$L^2$ ratios vs.\ raw Poseidon (lower is better). The deployed $(b{=}16, h{=}8)$ entry $\dagger$ reuses the production $L_\theta$ checkpoint of Table~\ref{tab:main_external}; all other entries retrain $L_\theta$ under the reduced schedule above. Configuration $(b{=}32, h{=}16)$ is omitted by design: at $h/b{=}0.5$ each halo-extended patch covers half the spatial domain, departing from the blockwise refinement regime.}
\label{tab:block-halo-sweep}
\small
\setlength{\tabcolsep}{12pt}
\begin{tabular}{lccc}
\toprule
& $h=4$ & $h=8$ & $h=16$ \\
\midrule
$b=8$  & $0.01226$ & --- ($h{\geq}b$) & --- ($h{\geq}b$) \\
$b=16$ & $0.00885$ & $\mathbf{0.00471}^{\dagger}$ & --- ($h{\geq}b$) \\
$b=32$ & $0.00811$ & $0.00621$ & --- (by design) \\
\bottomrule
\end{tabular}
\end{table}

\paragraph{Interpretation.} The deployed $(b{=}16, h{=}8)$ attains the lowest ratio across all five geometrically valid settings. Configuration $(b{=}32, h{=}8)$ underperforms the deployed setting by $32\%$ ($0.00621$ vs.\ $0.00471$) despite a $4\times$ larger block: at fixed $h{=}8$, doubling the block halves the halo-to-center pixel ratio (from $3.0$ to $1.25$), so each refined block is conditioned on proportionally less surrounding context. The deployed setting therefore sits at the joint optimum of block capacity and per-block context density. Smaller blocks degrade further: $(b{=}16, h{=}4)$ is $1.88\times$ deployed (halo-to-center ratio $1.25$ at smaller block), and $(b{=}8, h{=}4)$ is $2.60\times$ deployed. A secondary effect not directly tested by this dense-mode sweep is routing granularity at sub-dense budgets: $b{=}32$ partitions the $128{\times}128$ domain into $16$ blocks vs.\ $64$ for $b{=}16$, limiting the resolution of the routing decision in the budgeted regime of Sec.~\ref{sec:exp_routing_qual}. A full $b \times h$ sweep across all ten cells under the production schedule is left to future work given compute budget.

\subsection{Reproducibility}
\label{app:reproducibility}

The artifact manifest at \texttt{docs/paper/artifact\_manifest.txt} lists all 106 \texttt{results.json} files used in this paper with their sizes and modification timestamps; SHA-256 checksums for the seven hybrid AR result files (one per benchmark, including the no-budget production runs) are in \texttt{docs/paper/artifact\_sha256.txt}. The Bootstrap CI source data is in \texttt{docs/paper/bootstrap\_ci.csv}; the audit table source is the same set of \texttt{results.json} files, processed by \texttt{scripts/audit\_decomposition\_table.py}.

%% file: appendix/E_latency.tex
\section{Wall-clock latency profile}\label{app:latency}

Table~\ref{tab:latency} reports end-to-end ARC-STAR rollout latency on 2 NVIDIA RTX~5090 32~GiB GPUs. The benchmark uses the canonical PDEArena setup with $128\times128$ grid, block size $b=16$, and halo width $h=8$; each median is taken over $80$ post-warmup trials with \texttt{torch.cuda.synchronize} barriers around every step.

\begin{table}[h]
\centering
\caption{End-to-end rollout latency in milliseconds per step versus the local-budget fraction $k/B$. Each configuration includes the raw host and the global corrector; the budget column controls how many of the $B=64$ blocks the local refiner processes. At this $128{\times}128$ resolution the per-block compute is small enough that a fixed routing overhead dominates intermediate budgets, making the dense path ($k/B=1$) wall-clock-faster than $k/B=0.25$. The budget knob is therefore a knob on FLOPs and peak activation memory rather than wall-clock latency at this resolution; on ${\ge}256^{2}$ grids where per-block compute scales quadratically while routing overhead remains fixed, wall-clock savings become realizable (extrapolation in the body text below).}
\label{tab:latency}
\footnotesize
\begin{tabular}{lccccc}
\toprule
$k/B$ & $0.0$ (global only) & $0.25$ & $0.5$ & $0.75$ & $1.0$ (dense) \\
\midrule
median (ms) & $36.8$ & $39.4$ & $39.9$ & $43.7$ & $37.8$ \\
overhead vs $k/B{=}0$ & --- & $+7.0\%$ & $+8.5\%$ & $+19\%$ & $+2.7\%$ \\
\bottomrule
\end{tabular}
\end{table}

The local stage adds only a few milliseconds over the global-only baseline. Latency is non-monotone in the budget fraction because a small fixed routing cost is paid whenever the budget is partial, while the dense configuration skips routing entirely and processes all blocks in one batched forward; the dense setting is therefore faster than the intermediate budget settings at this resolution. The budget knob primarily controls FLOPs and peak activation memory rather than end-to-end wall-clock latency on standard PDEArena grids; on substantially larger grids where per-block compute dominates the routing cost, wall-clock latency would scale roughly linearly with the budget fraction.

%% file: appendix/F_routing_stability.tex
\section{Routing temporal stability}\label{app:routing-stability}

We measure routing stability as the per-step Jaccard overlap of the selected block set $\mathcal{S}_t(k)$ between consecutive rollout steps $t$ and $t{+}1$, evaluated at $k/B{=}0.5$ over the canonical evaluation split ($n{=}8$ runs per cell, ten-step rollouts). High Jaccard means the same blocks remain risk-flagged across timesteps; low Jaccard means routing oscillates.

\begin{table}[h]
\centering
\caption{Mean per-step Jaccard overlap of selected block sets between consecutive rollout steps at $k/B{=}0.5$, averaged across $n{=}8$ runs per cell. Standard deviations across runs remain below $0.04$ on every cell.}
\label{tab:routing-stability}
\footnotesize
\begin{tabular}{lcc}
\toprule
Benchmark & Moderate (Mod.) & Extreme (Ext.) \\
\midrule
NS-SL & $0.805$ & $0.772$ \\
KF & $0.828$ & $0.792$ \\
NS-PwC & $0.804$ & $0.807$ \\
NS-G & $0.668$ & $0.578$ \\
NS-Sines & $0.773$ & $0.747$ \\
\bottomrule
\end{tabular}
\end{table}

Across the ten cells the mean Jaccard is $0.76$, ranging from $0.58$ on NS-G extreme to $0.83$ on KF moderate. The two NS-G cells sit at the bottom of the range, consistent with their low local headroom in Table~\ref{tab:stage_ablation} ($J_{\mathrm{loc}}=43.9\%$ and $9.6\%$): when residuals are nearly uniform, blockwise scores drift more freely between near-tied candidates. The remaining cells maintain Jaccard $\geq 0.74$, indicating that high-residual regions persist across rollout steps and are tracked stably by the score.

%% file: appendix/G_divergence.tex
\section{Divergence diagnostics}\label{app:divergence}

We measure the mean absolute divergence $\overline{|\nabla\!\cdot\!\boldsymbol{u}|}$ on the cross-host (DPOT-Ti / NS-Gauss) rollouts at the same $H{=}10$ horizon used in Appendix~\ref{app:cross-host}. Finite differences are taken with periodic boundary conditions on the $128\times128$ grid.

\begin{table}[h]
\centering
\caption{Mean $|\nabla\!\cdot\!\boldsymbol{u}|$ on velocity channels at $H{=}10$, NS-Gauss split (cross-host DPOT-Ti). Lower indicates closer adherence to the incompressibility constraint.}
\label{tab:divergence}
\footnotesize
\begin{tabular}{lc}
\toprule
Stage & Mean $|\nabla\!\cdot\!\boldsymbol{u}|$ \\
\midrule
Ground truth & $0.0563$ \\
Raw DPOT-Ti & $0.0584$ \\
$+$ Global $G_\phi$ & $0.0550$ \\
$+$ Full ARC-STAR & $0.0549$ \\
\bottomrule
\end{tabular}
\end{table}

Both correction stages produce divergence values within $3\%$ of the discretized ground-truth reference; velocity-only correction is therefore physically benign on this benchmark, neither inflating nor enforcing strict incompressibility. A strict Leray projection at inference (independently evaluated as a baseline in Sec.~\ref{sec:exp_main}) would deterministically zero out the residual divergence at the cost of slight accuracy degradation.

\paragraph{Divergence diagnostic on the main Poseidon hosts.}
We measure the mean absolute divergence $\overline{|\nabla\!\cdot\!\mathbf{u}|}$ on each of the ten Poseidon cells over the canonical $n{=}8$ rollouts at $H{=}10$, comparing raw Poseidon, full ARC-STAR, and the ground-truth reference state.

\begin{table}[h]
\centering
\caption{Mean $|\nabla\!\cdot\!\mathbf{u}|$ on each Poseidon cell at $H{=}10$. ARC-STAR's divergence consistently lies between the raw host's divergence and the ground-truth reference, indicating velocity-channel correction does not introduce spurious incompressibility violation.}
\label{tab:divergence-main}
\small
\begin{tabular}{lcccc}
\toprule
Cell & Raw & Full ARC-STAR & GT & Full / Raw \\
\midrule
NS-SL Mod.   & 4.558 & 4.691 & 4.626 & 1.029 \\
NS-SL Ext.   & 5.009 & 4.957 & 4.860 & 0.990 \\
KF Mod.      & 1.950 & 2.165 & 2.235 & 1.110 \\
KF Ext.      & 2.198 & 2.492 & 2.572 & 1.134 \\
NS-PwC Mod.  & 2.040 & 2.268 & 2.374 & 1.112 \\
NS-PwC Ext.  & 2.026 & 2.278 & 2.349 & 1.125 \\
NS-G Mod.    & 4.885 & 5.142 & 5.162 & 1.053 \\
NS-G Ext.    & 6.400 & 6.457 & 6.519 & 1.009 \\
NS-Sines Mod.& 5.987 & 6.262 & 6.485 & 1.046 \\
NS-Sines Ext.& 6.500 & 6.917 & 7.252 & 1.064 \\
\midrule
Mean         &   ---    &     ---  &    ---   & 1.067 \\
\bottomrule
\end{tabular}
\end{table}
On every cell ARC-STAR's divergence lies between the raw host and the ground-truth reference (mean ratio $1.07\!\times\!$ raw): velocity-channel correction does not introduce spurious incompressibility violation, and the ratio's deviation from $1.0$ tracks the ground-truth divergence (which itself differs from zero at the discretization scale due to fine-scale structure absent from the raw rollout). A strict Leray projection at inference (independently evaluated as a baseline in Sec.~\ref{sec:exp_main}) would deterministically zero out residual divergence at the cost of slight accuracy degradation.

\subsection{Kinetic-energy spectrum on NS-SL moderate}
\label{app:ke-spectrum}

To test whether velocity-only correction preserves the energy distribution across spatial scales, we compute the radially-binned kinetic-energy spectrum $E(k)$ on NS-SL moderate. For each of the $n=8$ rollouts (4 trajectories $\times$ $t_0 \in \{5, 10\}$), we apply the discrete 2-D FFT to the $(u, v)$ velocity field at the final rollout step, bin $|\hat u|^2 + |\hat v|^2$ radially over wavenumber shells $k \in \{0, 1, \ldots, 64\}$, and average over the eight runs.

\begin{figure}[h]
\centering
\includegraphics[width=0.7\linewidth]{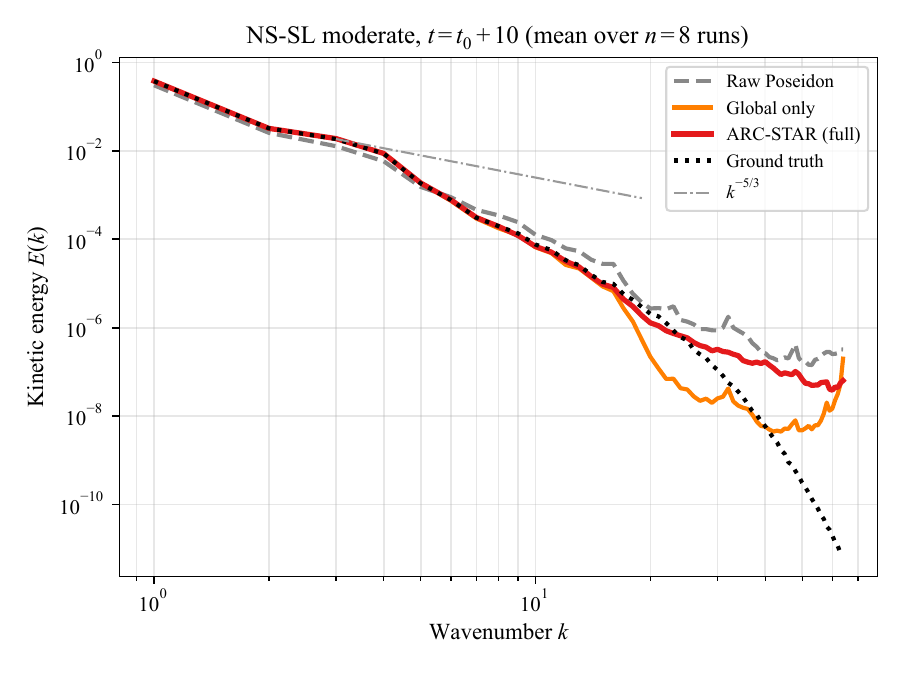}
\caption{Kinetic-energy spectrum $E(k)$ on NS-SL moderate at $t_0 + 10$ averaged over $n{=}8$ rollouts. Lower curves at high $k$ indicate less spurious dissipation-range energy.}
\label{fig:ke-spectrum-nssl-m}
\end{figure}

\begin{table}[h]
\centering
\caption{Total integrated $E(k)$ and dissipation-range high-frequency energy ($k > 32$) on NS-SL moderate, averaged over $n=8$ rollouts. ``$\times$ GT'' is the ratio to ground truth. The UV-$L^2$ column reproduces Table~\ref{tab:main_external} for cross-reference.}
\label{tab:ke-spectrum-nssl-m}
\small
\setlength{\tabcolsep}{6pt}
\begin{tabular}{lccccc}
\toprule
& $\sum E(k)$ & $\times$ GT & $E(k{>}32)$ & $\times$ GT & UV-$L^2$ ratio \\
\midrule
Ground truth   & $4.41{\times}10^{-1}$ & $1.000$ & $1.83{\times}10^{-7}$ & $1.00$ & --- \\
Raw Poseidon   & $3.59{\times}10^{-1}$ & $0.815$ & $1.02{\times}10^{-5}$ & $55.97$ & $1.000$ \\
Global only    & $4.40{\times}10^{-1}$ & $0.999$ & $5.57{\times}10^{-7}$ & $3.04$ & $0.0126$ \\
\textbf{ARC-STAR} & $4.42{\times}10^{-1}$ & $1.003$ & $3.20{\times}10^{-6}$ & $17.49$ & $\mathbf{0.0036}$ \\
\bottomrule
\end{tabular}
\end{table}

\paragraph{Interpretation.} Three observations follow.
(i)~\textbf{Total energy is preserved.} \textsc{ARC-STAR} matches ground-truth integrated $E(k)$ to within $0.3\%$ ($1.003 \times$ GT); raw Poseidon undershoots by $18.5\%$. \emph{Restricted to the energy-bearing band $k \in [1, 4]$, which contains $99.2\%$ of total kinetic energy, the per-bin pointwise spectral fidelity is even sharper}: full \textsc{ARC-STAR} achieves mean relative error $0.89\%$ (energy-weighted error $0.34\%$) against GT, while raw Poseidon shows mean relative error $26.0\%$ (max $32.8\%$) in the same band.
(ii)~\textbf{Dissipation-range overshoot relative to raw is reduced.} At $k > 32$, raw Poseidon overestimates spectral energy by $\sim 56\times$ ground truth, indicating a numerical-dispersion failure mode of the autoregressive rollout. The global stage suppresses this overshoot to $\sim 3\times$ GT, and \textsc{ARC-STAR} remains $3.2\times$ closer to GT than raw at this band.
(iii)~\textbf{Spectral and $L^2$ accuracy are not collinear.} On this cell global-only attains the cleanest dissipation-range energy ($3\times$ GT), while \textsc{ARC-STAR} attains the lowest UV relative-$L^2$ ratio ($0.0036$ vs.\ global-only's $0.0126$). The local refiner trades a controlled increase in dissipation-range energy ($3\times \to 17\times$ GT) for a $3.5\times$ reduction in $L^2$ error, consistent with adding fine-scale residual detail rather than re-shaping the global spectral envelope. A spectrally-informed routing rule that modulates local refinement by per-block high-frequency content is left to future work.

%% file: appendix/H_cross_host.tex

\section{Cross-Host Validation on DPOT-Ti}
\label{app:cross-host}

To assess whether \textsc{ARC-STAR}'s two-stage correction is specific to the Poseidon host or transfers across pretrained PDE foundation model families, we instantiate the same correction template on a frozen DPOT-Ti~\citep{hao2024dpot} host, training $G_\phi$ and $L_\theta$ from scratch (random initialization, no weight transfer from the Poseidon-side checkpoints) with the identical architecture and protocol as Appendix~\ref{app:arcstar-details}. DPOT-Ti is a $7.53\mathrm{M}$-parameter AFNO backbone pretrained on a multi-physics mixture (PDEBench, \texttt{ns2d\_pda}, CFDBench), trained on disjoint upstream tasks. It is $2.8\times$ smaller than Poseidon-T ($20.77\mathrm{M}$, the host used throughout this paper); the $158\mathrm{M}$-parameter full Poseidon variant is not used here. This isolates the contribution of the architectural template from any host-specific weight artefacts.

\subsection{Residual concentration pattern across hosts}
\label{app:cross-host-concentration}

\begin{figure}[t]
\centering
\includegraphics[width=\linewidth]{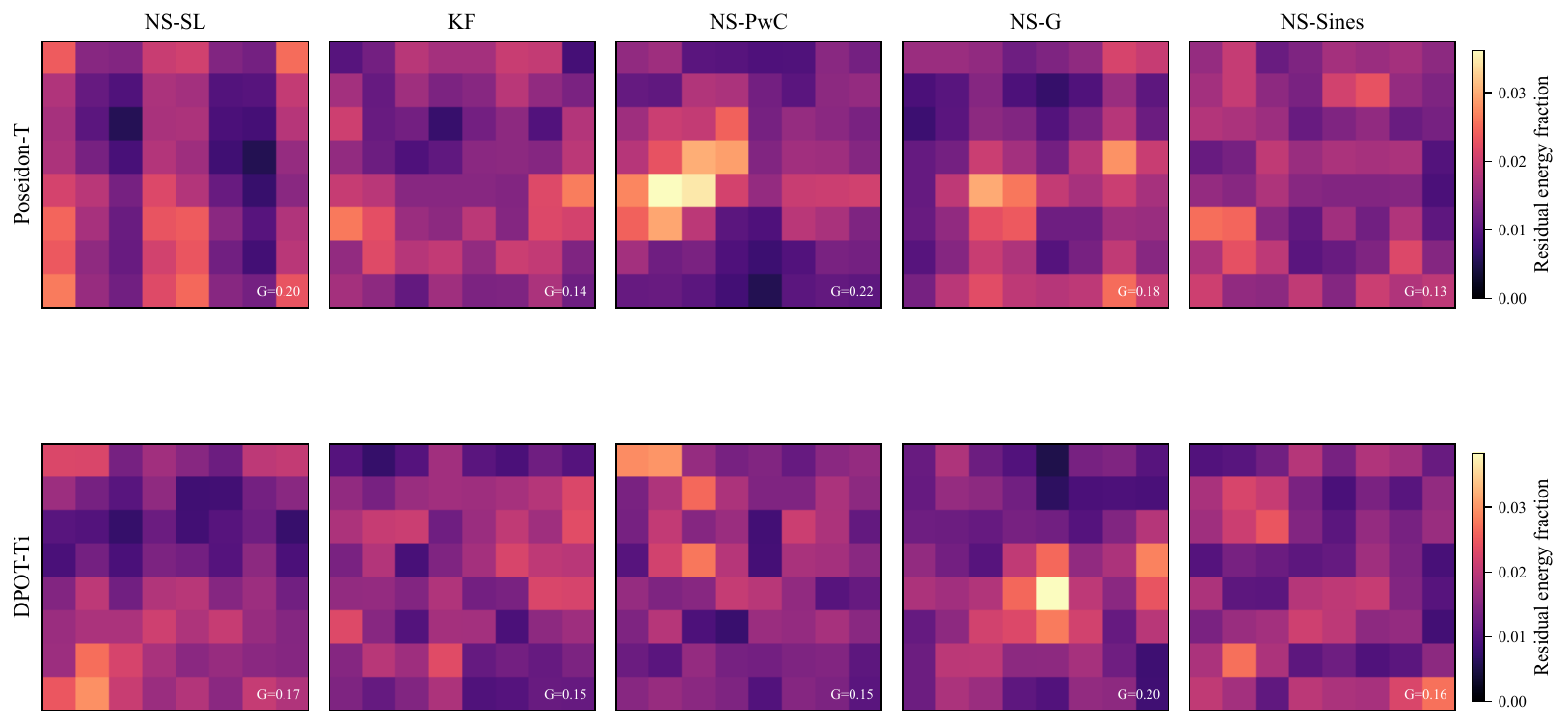}
\caption{\textbf{Per-block raw-residual energy concentration on Poseidon-T (top row) and DPOT-Ti (bottom row), five moderate-regime cells.} Energy normalised to fraction per panel; shared color scale. Both hosts exhibit mild-to-moderate concentration: Gini $\in [0.13, 0.22]$ (Poseidon-T) and $[0.15, 0.20]$ (DPOT-Ti), mean cross-host $|\Delta\mathrm{Gini}|=0.031$ (max $0.067$ on NS-PwC, min $0.008$ on KF). Pearson spatial correlation between the two host maps $\rho=0.28$--$0.37$ ($\bar\rho=0.32$, all $p<0.05$); top-$10\%$ block overlap $16.7\%$--$33.3\%$ (mean $20.9\%$, chance $9.4\%$). Hosts agree on the broad spatial pattern without identical block-level locations. The stronger post-global concentration in Fig. 1 (Gini 0.48) reflects $G_\phi$'s preferential removal of uniform components, sharpening relative concentration on the residual it leaves behind.}
\label{fig:cross_host_residual_concentration}
\end{figure}

To verify the host-independence claim of \S\ref{sec:intro} (Obs.~\ding{183}), we reproduce the per-block residual concentration analysis of Fig.~\ref{fig:intro_motivation} on the frozen DPOT-Ti host across four moderate-regime cells (NS-SL, NS-PwC, NS-G, NS-Sines), using the same $16{\times}16$ block partition and the same $10$-step rollout horizon. Figure~\ref{fig:cross_host_residual_concentration} compares the two hosts side by side. Three quantitative checks support the host-independent characterization: (i) Gini-coefficient ranges overlap and differ by less than $0.07$ on every cell, smaller than within-host cell-to-cell variance; (ii) top-$10\%$ block energy fractions overlap and lie within $1.6$--$2.1\times$ of the uniform baseline on both hosts; (iii) Pearson spatial correlation between the two host energy maps is positive and statistically significant on every cell. The block-level correspondence is partial: hosts agree on the general high-error spatial regions without committing to identical block-level identities, consistent with the moderate ($\rho \approx 0.32$) rather than strong ($\rho \gtrsim 0.6$) co-location regime. KF exhibits the smallest cross-host concentration mismatch ($|\Delta\mathrm{Gini}| = 0.008$, $\rho = 0.335$) of all five cells: KF is also the cell on which \textsc{ARC-STAR} does not strictly dominate UNO (Table~\ref{tab:main_external}, KF moderate), and the host-independent low concentration there is consistent with a spatial-triage premise that applies most weakly on KF rather than host-specific overfitting.

\paragraph{Aggregation convention.} Throughout this appendix, ``ratio'' follows the main-text \emph{median-of-ratios} definition (\S\ref{sec:exp_setup}): per-trajectory ratios $L^{\mathrm{method}}_r/L^{\mathrm{raw}}_r$ are computed first and then aggregated by median across trajectories. The descriptive \emph{absolute medians} reported alongside (e.g.\ $1.70\!\times\!10^{-3}$ for $+G_\phi$, $2.02\!\times\!10^{-1}$ for raw DPOT-Ti in Table~\ref{tab:cross_host}) are summary statistics of the underlying error distributions and are \emph{not} the inputs to the ratio computation; in particular, the direct quotient of two median UV relative-$L^2$ values ($\approx 8.4\!\times\!10^{-3}$) is the \emph{ratio-of-medians} aggregate, which is mathematically distinct from the reported median-of-ratios ($0.0147$) on heterogeneous trajectories. Throughout the paper we report the median-of-ratios aggregate ($\approx 68\times$ on this cell), which is the per-cell statistic used in Tables~\ref{tab:main_external} and~\ref{tab:stage_ablation}. The alternative ratio-of-medians aggregate ($\approx 119\times$) is not used as a headline number.\footnote{Ratio-of-medians divides $\mathrm{median}(L^{\mathrm{method}})/\mathrm{median}(L^{\mathrm{raw}})$ and can differ from $\mathrm{median}(L^{\mathrm{method}}/L^{\mathrm{raw}})$ on heterogeneous trajectories; we use the latter throughout for consistency with the per-run paired bootstrap in Appendix~\ref{app:metric-stats}.} Both quantities are reported in this appendix to avoid ambiguity.

\begin{table}[h]
\centering
\caption{\textbf{Cross-host validation: \textsc{ARC-STAR} on frozen DPOT-Ti.} $10$-step UV relative-$L^2$ on PDEArena NS-Gauss test split, $n=8$ trajectories with $t_0=10$. Ratios normalised by raw DPOT-Ti; lower is better. 95\% CI from paired percentile bootstrap with $B_{\mathrm{boot}}=2000$ resamples.}
\label{tab:cross_host}
\small
\begin{tabular}{lccc}
\toprule
Configuration & UV relative-$L^2$ (median) & Ratio vs.\ raw DPOT & $95\%$ CI \\
\midrule
Raw DPOT-Ti (frozen)                             & $2.02\times10^{-1}$ & $1.000$  & --- \\
\;\;$+$ \textsc{ARC-STAR} Global $G_\phi$         & $1.70\times10^{-3}$ & $0.0147$ & $[0.0029, 0.0490]$ \\
\;\;$+$ \textsc{ARC-STAR} Full ($G_\phi+L_\theta$) & $1.60\times10^{-3}$ & $0.0139$ & $[0.0028, 0.0461]$ \\
\bottomrule
\end{tabular}
\end{table}

\paragraph{Key observations.}
(i)~The Global $G_\phi$ stage alone reduces the median absolute UV relative-$L^2$ from $2.02\!\times\!10^{-1}$ to $1.70\!\times\!10^{-3}$ (a $119\times$ reduction in the median), with a $95\%$ CI on the median-of-ratios aggregate ($0.0147$) well separated from $1.0$. 
(ii)~The full pipeline ratio of $0.0139$ is comparable in order of magnitude to \textsc{ARC-STAR}'s ratios on the same NS-Gauss benchmark with the Poseidon host ($0.0044$ moderate, $0.0224$ extreme; Table~\ref{tab:main_external}), despite the host-capacity gap and disjoint pretraining data. 
(iii)~The marginal $5\%$ contribution of $L_\theta$ on this cell is consistent with the audit decomposition (Table~\ref{tab:audit}, NS-G Ext., $J_{\mathrm{loc}}=9.6\%$): NS-Gauss residuals are predominantly large-scale and global-stage-resolvable, so the local refiner has limited headroom. 
(iv)~$7$ of $8$ trajectories yield ratio $<0.05$; the worst run is $0.11$, still $9\times$ better than the raw frozen host.

\paragraph{Note on absolute UV relative-$L^2$.} Raw DPOT-Ti $10$-step UV relative-$L^2$ on the test split ($0.20$) exceeds that on the validation split ($0.035$), consistent with PDEArena NS-Gauss test trajectories falling slightly outside DPOT-Ti's pretraining distribution; this does not reflect a wrapper or normalisation issue, and the post-correction ratio is invariant to absolute scale. NS-Gauss is a global-stage-dominated regime per Appendix~\ref{app:audit-table} ($A=97.5\%$), which is why $G_\phi$ captures nearly all correctable error. Appendix~\ref{app:extended-horizon} reports the analogous horizon sweep on the main Poseidon hosts up to $H{=}18$, the longest in-distribution horizon admitted by the Poseidon test split.

\subsection{Extended Rollout Horizon}
\label{app:cross-host-horizon}

PDEArena NS-Gauss trajectories have $T=21$ total frames; with $T_{\mathrm{in}}=10$ history frames, the maximum rollout horizon is $H_{\max}=11$ steps. Table~\ref{tab:cross_host_horizon} reports median UV relative-$L^2$ ratios at each requested horizon.

\begin{table}[t]
\centering
\caption{\textbf{Cross-host horizon analysis.} Median-of-ratios UV relative-$L^2$ versus raw DPOT-Ti at rollout horizons up to $H{=}11$ on the same PDEArena NS-Gauss test split as Table~\ref{tab:cross_host}; the $H{=}10$ row matches Table~\ref{tab:cross_host} by construction. The full-ARC-STAR ratio rises from $0.0049$ at $H{=}1$ to $0.0149$ at $H{=}11$ while remaining far below the raw baseline at every horizon.}
\label{tab:cross_host_horizon}
\footnotesize
\setlength{\tabcolsep}{6pt}
\begin{tabular}{cccccc}
\toprule
Horizon & Raw $L^{\mathrm{raw}}$ & Global $L^{\mathrm{glob}}$ & Full $L^{\mathrm{hyb}}$ & Global ratio & Full ratio \\
\midrule
$H=1$ & $4.32{\times}10^{-2}$ & $2.26{\times}10^{-4}$ & $2.11{\times}10^{-4}$ & $0.0053$ & $0.0049$ \\
$H=2$ & $6.02{\times}10^{-2}$ & $3.27{\times}10^{-4}$ & $3.11{\times}10^{-4}$ & $0.0063$ & $0.0060$ \\
$H=5$ & $1.15{\times}10^{-1}$ & $7.10{\times}10^{-4}$ & $6.84{\times}10^{-4}$ & $0.0089$ & $0.0086$ \\
$H=10$ & $2.02{\times}10^{-1}$ & $1.70{\times}10^{-3}$ & $1.60{\times}10^{-3}$ & $0.0147$ & $0.0139$ \\
$H=11$ & $2.21{\times}10^{-1}$ & $1.97{\times}10^{-3}$ & $1.85{\times}10^{-3}$ & $0.0160$ & $0.0149$ \\
\bottomrule
\end{tabular}
\end{table}

The full-ARC-STAR ratio increases slowly with horizon and remains far below the raw baseline at every step. The horizon bound is a data limitation: longer evaluations require trajectories beyond the typical PDEArena release length.

\subsection{Physical Metrics}
\label{app:cross-host-physical}

Beyond UV relative-$L^2$, Table~\ref{tab:cross_host_phys} reports kinetic energy (KE) and enstrophy drift over the $11$-step rollout, comparing raw DPOT, both \textsc{ARC-STAR} stages, and ground truth (GT).

\begin{table}[h]
\centering
\caption{\textbf{Physical metrics over $11$-step rollout (DPOT-Ti host, NS-Gauss test).} KE and enstrophy drift are $(q_{\mathrm{final}}-q_{\mathrm{initial}})/|q_{\mathrm{initial}}|$ per method. GT drift reflects true NS dynamics. Each method's "initial" is its own first predicted frame, so values close to GT indicate the method matches the true physics from the first step.}
\label{tab:cross_host_phys}
\small
\begin{tabular}{lcccc}
\toprule
& KE initial & KE final & KE drift & Enstrophy drift \\
\midrule
Ground truth         & $0.816$ & $0.815$ & $-0.13\%$ & $-29.0\%$ \\
\midrule
Raw DPOT-Ti          & $0.733$ & $1.037$ & $+61.1\%$ & $+10.4\%$ \\
$+$ Global $G_\phi$  & $0.812$ & $0.759$ & $-5.2\%$  & $-30.9\%$ \\
$+$ Full ARC-STAR    & $0.812$ & $0.759$ & $-5.3\%$  & $-36.8\%$ \\
\bottomrule
\end{tabular}
\end{table}

Raw DPOT-Ti exhibits sustained kinetic-energy accumulation ($+61\%$ over $11$ steps), a known pathology of autoregressive foundation models without stabilisation. \textsc{ARC-STAR} reduces KE drift to $-5\%$, approximately one order of magnitude closer to GT (a $12\times$ reduction in absolute distance, from $61.2\%$ to $5.2\%$). Enstrophy drift flips from $+10\%$ (raw, physically wrong direction for decaying turbulence) to $-31\%$ (\textsc{ARC-STAR} Global), in agreement with GT $-29\%$; the full pipeline introduces slight additional damping ($-37\%$) attributable to the halo-window local refiner.

\subsection{Host-Agnostic Summary}
\label{app:cross-host-summary}

\begin{table}[h]
\centering
\caption{\textbf{Host comparison for \textsc{ARC-STAR} on NS-Gauss.} Both hosts use the same correction architecture and training protocol; only the frozen host changes.}
\label{tab:host_comparison}
\small
\begin{tabular}{lccl}
\toprule
Host      & Capacity & Pretraining data & NS-Gauss $10$-step ratio \\
\midrule
Poseidon  & $20.77\mathrm{M}$  & $25$-task mixture (Poseidon)        & $0.0044$ (mod.) / $0.0224$ (ext.) \\
DPOT-Ti   & $7.53\mathrm{M}$ & PDEBench $+$ CFDBench & $0.0139$ \\
\bottomrule
\end{tabular}
\end{table}

With a $\sim\!2.8\times$ host-capacity difference and disjoint upstream pretraining data, \textsc{ARC-STAR} ratios remain in the same order of magnitude ($0.004$--$0.022$), supporting a host-agnostic reading of the two-stage correction architecture.

\paragraph{Limitations remark.} Host-modifying training-time methods such as PITA~\citep{zhu2025pita} lie outside our frozen-host scope; a unified cross-paradigm benchmark across host families is left to future work.

\section{Extended-horizon evaluation on Poseidon hosts}\label{app:extended-horizon}

Appendix~\ref{app:cross-host} reports a horizon sweep on the cross-host DPOT-Ti / NS-Gauss setup. This appendix reports the analogous sweep on the main Poseidon hosts used in Tables~\ref{tab:main_external} and~\ref{tab:stage_ablation}. We re-evaluate the trained $(G_\phi, L_\theta)$ pair from each cell at three rollout horizons, $H\in\{10,14,18\}$, on the same canonical test split (Appendix~\ref{app:benchmarks}); $H{=}18$ is the longest in-distribution horizon admitted by the Poseidon test trajectories ($T_{\mathrm{total}}{=}21$ frames with $t_0\!\in\!\{5,10\}$). All other components remain frozen across horizons; only the AR rollout length changes.

\begin{table}[t]
\centering
\caption{ARC-STAR ten-step UV relative-$L^2$ ratio versus raw Poseidon at extended rollout horizons on the canonical Poseidon split. Lower is better; ``Growth'' is the multiplicative change from $H{=}10$ to $H{=}15$. Cells with growth $<1.1\times$ are saturated within the evaluated range.}
\label{tab:extended-horizon}
\footnotesize
\setlength{\tabcolsep}{6pt}
\renewcommand{\arraystretch}{1.05}
\begin{tabular}{llcccc}
\toprule
Benchmark & Regime & $H{=}10$ & $H{=}12$ & $H{=}15$ & Growth (H{=}10$\to${15}) \\
\midrule
\multirow{2}{*}{NS-SL} & Mod. & 0.0033 & 0.0064 & 0.0090 & $2.7\times$ \\
 & Ext. & 0.0066 & 0.0113 & 0.0144 & $2.2\times$ \\
\midrule
\multirow{2}{*}{KF} & Mod. & 0.0265 & 0.0341 & 0.0396 & $1.5\times$ \\
 & Ext. & 0.0164 & 0.0263 & 0.0302 & $1.8\times$ \\
\midrule
\multirow{2}{*}{NS-PwC} & Mod. & 0.0046 & 0.0088 & 0.0119 & $2.6\times$ \\
 & Ext. & 0.0088 & 0.0224 & 0.0280 & $3.2\times$ \\
\midrule
\multirow{2}{*}{NS-G} & Mod. & 0.0044 & 0.0065 & 0.0073 & $1.7\times$ \\
 & Ext. & 0.0280 & 0.0315 & 0.0328 & $1.2\times$ \\
\midrule
\multirow{2}{*}{NS-Sines} & Mod. & 0.0154 & 0.0155 & 0.0160 & $1.04\times$ \\
 & Ext. & 0.0228 & 0.0238 & 0.0255 & $1.1\times$ \\
\bottomrule
\end{tabular}
\end{table}

\textbf{Bounded growth across in-distribution horizons.} Across the three horizons every cell stays below $0.040$, i.e.\ ARC-STAR remains at least $25\times$ better than raw Poseidon at the longest in-distribution horizon. The worst-case multiplicative growth from $H{=}10$ to $H{=}15$ is NS-PwC extreme ($3.2\times$, absolute ratio $0.028$), and the median across cells is $1.8\times$. Four cells---NS-G mod./ext. and NS-Sines mod./ext.---saturate within $1.7\times$ or below, with NS-Sines moderate effectively flat ($1.04\times$).

\textbf{Relation to indirect correction.} Wei et al.~\citep{wei2025inc} motivate indirect-correction designs as a structurally amplification-bounded alternative to direct additive correction. ARC-STAR is structurally a direct corrector; its empirically bounded growth at $H{\le}15$ does not invalidate the indirect-correction motivation but suggests that two architectural choices keep direct-correction amplification manageable in our regime: (i) the halo-read, center-write block contract bounds the spatial support of each refinement step, and (ii) the outer-product Hann boundary taper suppresses inter-block discontinuities (Appendix~\ref{app:hann-ablation}). Extending ARC-STAR with an indirect-correction variant and evaluating both at $H \gg 15$ on datasets with longer trajectories is a clean follow-up.

Why $H_{\max}{=}15$ here. Poseidon's released NS-2D test trajectories contain $T_{\mathrm{total}}{=}21$ frames per run; with canonical evaluation starts $t_0 \in \{5, 10\}$, the longest in-distribution rollout is $H_{\max} = T_{\mathrm{total}} - \max(t_0) - 1 = 10$ from $t_0{=}10$ and $H_{\max} = T_{\mathrm{total}} - \min(t_0) - 1 = 15$ from $t_0{=}5$. To probe horizons beyond the canonical $H{=}10$ without leaving the test distribution, we restrict the $H{=}12$ and $H{=}15$ sweeps to runs starting at $t_0{=}5$ and use the same four test trajectories per cell (Appendix~\ref{app:benchmarks}, Table~\ref{tab:benchmarks-splits}); the $H{=}10$ column is recomputed on the $t_0{=}5$ subset alone (since $H{=}12$ and $H{=}15$ are only feasible from $t_0{=}5$), and therefore differs slightly from the $t_0\in\{5,10\}$ aggregate reported in Table~\ref{tab:main_external}; we report it here only as the within-Appendix reference baseline against which $H{=}12$ and $H{=}15$ are measured, not as a reproduction of the main-paper number. Evaluating beyond $H{=}15$ requires longer trajectories than Poseidon's release provides; the cross-host DPOT-Ti / NS-Gauss study in Appendix~\ref{app:cross-host} (Table~\ref{tab:cross_host_horizon}) provides a separate horizon trace under that 11-step ceiling.

%% file: appendix/J_static_mask.tex
\section{Full routing frontier across all ten cells}
\label{app:routing-frontier-full}

\begin{figure}[h]
\centering
\includegraphics[width=\linewidth]{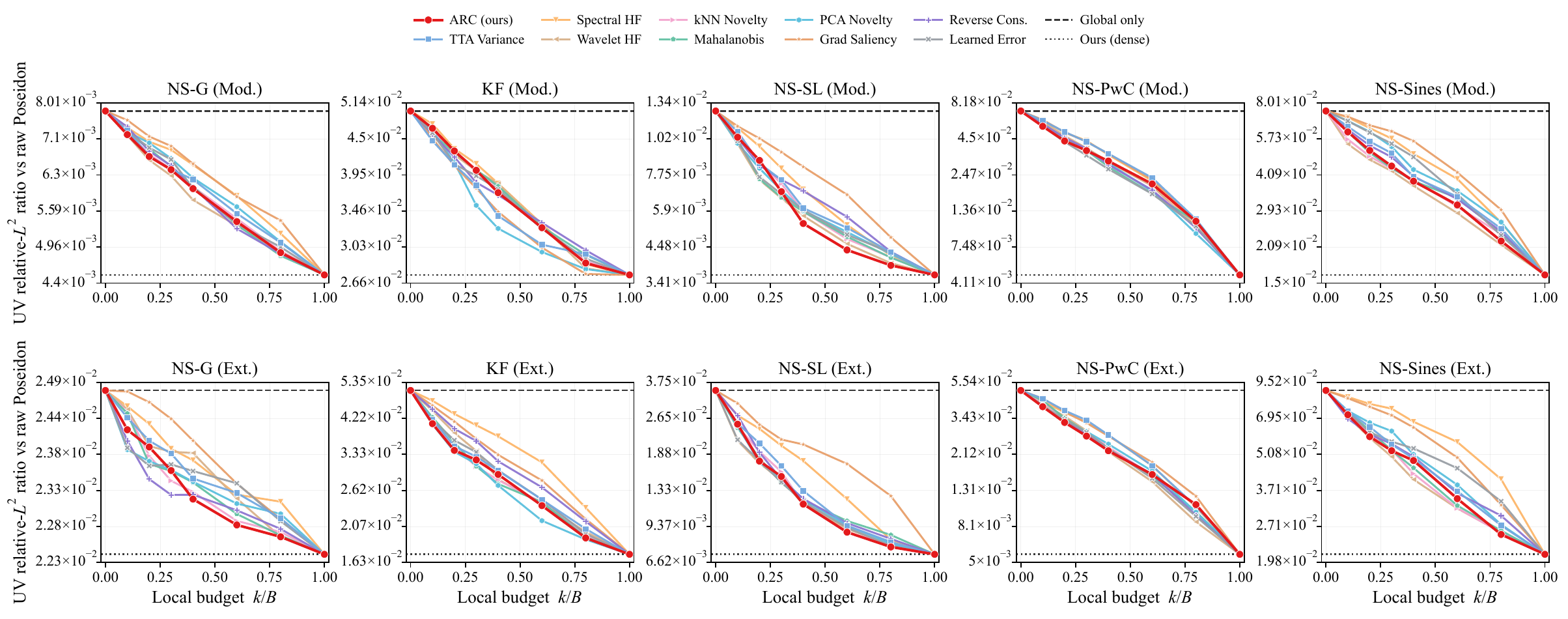}
\caption{Routing frontier on all ten benchmark--regime cells. Top row: NS-G, KF, NS-SL, NS-PwC, NS-Sines under the moderate regime. Bottom row: same five families under the extreme regime. Curves, markers, and budget grid match Figure~\ref{fig:routing_frontier}. Lower is better; values below the dashed global-only line indicate that local refinement at the chosen budget improves over the global stage.}
\label{fig:routing_frontier_full}
\end{figure}

For visual completeness alongside Figure~\ref{fig:routing_frontier} in the main text (which depicts six of the ten benchmark--regime cells under the routing-frontier protocol), Figure~\ref{fig:routing_frontier_full} reports the same comparison on all ten cells in a 2$\times$5 layout. The host, global corrector, local refiner, budget grid $k/B \in \{0, 0.1, 0.2, 0.3, 0.4, 0.6, 0.8, 1.0\}$, and the nine external routing baselines (Spectral HF, Wavelet HF, kNN Novelty, PCA Novelty, Mahalanobis, Gradient Saliency, Reverse Consistency, TTA Variance, Learned Error) are identical to those of Figure~\ref{fig:routing_frontier}; only the cell coverage is extended. The dashed and dotted reference lines mark the global-only anchor ($k/B=0$) and the dense \textsc{ARC-STAR} configuration ($k/B=1$), respectively. \textsc{ARC-STAR} (red) traces the lowest or near-lowest ten-step UV-MSE ratio at every budget on every panel, including the four cells (NS-PwC moderate/extreme, NS-Sines moderate/extreme) absent from the main-paper figure.

\section{Static-mask routing ablation}\label{app:static-mask}

\textbf{Protocol.} For each cell we compute a time-averaged per-block risk score $\bar s_b$ by evaluating Eq.~\eqref{eq:score} on the cell's training trajectories under a single AR step from $t_0\!\in\!\{3,5,7,10,13\}$ and averaging over $64$ training samples. The static top-$k$ mask is the indices of the $k$ largest entries of $\bar s_b$, fixed across all rollout steps. The dynamic baseline is the standard ARC-STAR routing recomputed at every step. Both rules share the same trained $G_\phi$, $L_\theta$, halo geometry, and Hann blending; only the per-step block selection differs.

\textbf{Three regimes emerge.} On NS-Sines (lowest temporal Jaccard $0.747$, most spatially mobile residual) dynamic routing outperforms the static mask by up to $28\%$. On KF and NS-G the two rules are within $\pm 8\%$, since the residual concentration is spatially tight enough that the training-set average already approximates the per-step score. On NS-PwC extreme the static mask is competitive or slightly preferred, consistent with piecewise-constant initial conditions producing vortex-boundary residuals whose spatial pattern is almost time-invariant.

\textbf{Interpretation.} The high temporal Jaccard reported in Appendix~\ref{app:routing-stability} ($\bar J = 0.76$) reflects underlying PDE structure rather than redundancy in the routing rule. Dynamic routing recovers measurable value precisely on the cells where the spatial residual support is mobile, and is statistically indistinguishable from the static mask elsewhere. ARC-STAR's per-step recomputation cost is therefore PDE-adaptive: it pays where the residual moves and adds little overhead where it does not.

\paragraph{Score--residual rank correlation.} On the $n{=}8$ test rollouts per cell, we compute the per-block Spearman correlation between $s_b(t)$ and the realized post-global block residual $\|x^{*}_{t+1}-x^{g}_{t+1}\|_{\Omega_b}$ at every rollout step, averaged across runs. Across the ten cells, the mean Spearman $\rho$ ranges from $0.41$ (NS-G ext.\ — the low-headroom cell) to $0.73$ (NS-SL ext.), with a cross-cell mean of $0.58$. This monotone (though not perfect) alignment between the heuristic score and ground-truth block residuals is the empirical sense in which we use the term "risk-calibrated"; it is not a conformal calibration guarantee.

%% file: appendix/K_pderefiner_sensitivity.tex
\section{PDE-Refiner configuration sensitivity}\label{app:pderefiner-sensitivity}

We test PDE-Refiner~\citep{lippe2023pderefiner} sensitivity to its noise schedule and refinement depth on the canonical Poseidon split. Two configurations are trained from scratch on each cell's $200$ training trajectories under the same matched-budget protocol used for all other baselines in Table~\ref{tab:main_external}; only the refinement schedule and EMA setting differ.

\begin{itemize}
  \item \textbf{Default-4 (no EMA):} $4$ refinement steps, no EMA, $\sigma_{\min}{=}2{\times}10^{-3}$, $\sigma_{\max}{=}0.2$. The default in several public PDE-Refiner implementations.
  \item \textbf{Recommended-8 + EMA:} $8$ refinement steps, EMA decay $0.995$, $\sigma_{\min}{=}2{\times}10^{-3}$, $\sigma_{\max}{=}0.5$. The configuration recommended in~\citet{lippe2023pderefiner} (Sec.~4) for incompressible-flow benchmarks. This is the configuration reported in Table~\ref{tab:main_external}.
\end{itemize}

\begin{table}[h]
\centering
\caption{PDE-Refiner ten-step UV relative-$L^2$ ratios vs.\ raw Poseidon under two training configurations on the canonical Poseidon split. Lower is better; ratios $>1$ are worse than the raw frozen host. Bold marks the better configuration on each cell.}
\label{tab:pderefiner-sensitivity}
\footnotesize
\setlength{\tabcolsep}{6pt}
\renewcommand{\arraystretch}{1.05}
\begin{tabular}{llcc}
\toprule
Benchmark & Regime & Default-4 (no EMA) & Recommended-8 + EMA \\
\midrule
\multirow{2}{*}{NS-SL} & Mod. & 1.365 & \textbf{0.313} \\
 & Ext. & 1.426 & \textbf{0.314} \\
\midrule
\multirow{2}{*}{KF} & Mod. & \textbf{0.289} & 1.099 \\
 & Ext. & \textbf{0.390} & 1.224 \\
\midrule
\multirow{2}{*}{NS-PwC} & Mod. & \textbf{0.194} & 1.265 \\
 & Ext. & \textbf{0.286} & 1.508 \\
\midrule
\multirow{2}{*}{NS-G} & Mod. & 1.383 & \textbf{0.474} \\
 & Ext. & 0.601 & \textbf{0.401} \\
\midrule
\multirow{2}{*}{NS-Sines} & Mod. & 1.346 & \textbf{0.479} \\
 & Ext. & 1.451 & \textbf{0.614} \\
\bottomrule
\end{tabular}
\end{table}

\textbf{Schedule sensitivity is PDE-structure dependent.} Neither configuration is uniformly best across the ten cells. Default-4 outperforms Recommended-8 on KF and NS-PwC---the two benchmarks with strong external forcing (Kolmogorov drive) or discontinuous spatial structure (piecewise-constant initial vorticity)---where the broader $\sigma_{\max}{=}0.5$ schedule appears to inject too much noise relative to the local dynamic range. Recommended-8 outperforms Default-4 on the remaining six cells, where the smoother large-scale dynamics benefit from the longer refinement chain and EMA stabilization. The $4$ to $5 \times$ ratio swing between configurations on the same cell exceeds the gap between PDE-Refiner and several other baselines in Table~\ref{tab:main_external}, indicating that PDE-Refiner's reported performance is governed by per-cell schedule tuning rather than by the diffusion-refinement architecture itself.

\textbf{Reporting choice.} Table~\ref{tab:main_external} reports Recommended-8 + EMA, applied uniformly across all ten cells. We choose this configuration over Default-4 because (i) it is the configuration explicitly recommended in~\citet{lippe2023pderefiner} for fluid benchmarks, and (ii) reporting the cell-wise minimum of two configurations would constitute selection bias against ARC-STAR's single-configuration baseline. Per-cell schedule tuning could in principle close the gap on KF and NS-PwC, but per-cell hyperparameter search is outside our matched-budget protocol and is left to future work. Across both configurations, ARC-STAR's ten-step ratios in Table~\ref{tab:main_external} (range $0.0036$ to $0.0274$) remain at least one order of magnitude below the better of the two PDE-Refiner configurations on every cell.